\documentclass{article}

\usepackage{microtype}
\usepackage{graphicx}
\graphicspath{{figures/figures_for_overleaf/}}
\usepackage[export]{adjustbox}

\usepackage{subcaption}
\usepackage{booktabs} 
\usepackage{float}
\usepackage{tcolorbox}
\usepackage{placeins}
\usepackage{url}
\usepackage{hyperref}

\usepackage{hyperref}

\usepackage[accepted]{icml2026_fogen}

\usepackage{amsmath}
\usepackage{amssymb}
\usepackage{mathtools}
\usepackage{amsthm}

\usepackage[capitalize,noabbrev]{cleveref}

\theoremstyle{plain}

\theoremstyle{definition}

\theoremstyle{remark}

\usepackage[textsize=tiny]{todonotes}

\icmltitlerunning{Interpreting Latent CoT Reasoning as Dynamical Systems}

\begin{document}

\noindent{\small ICML 2026 Workshop on Foundations of Deep Generative Models:
Understanding Memorization, Generalization, and Reasoning}
\vskip 0.35in

\icmltitle{Interpreting Latent CoT Reasoning as Dynamical Systems}


\icmlsetsymbol{equal}{*}
\icmlsetsymbol{lead}{$\dagger$}

\begin{icmlauthorlist}
  \icmlauthor{Sabari Iyyappan Duraipandian}{equal,sjsu}
  \icmlauthor{Shreya Sanjay Boyane}{equal,wpi}
  \icmlauthor{Manju Nagesh}{gmu}
  \icmlauthor{Jerome Francis}{lead,algoverse}
  \icmlauthor{Archana Vaidheeswaran}{algoverse}
  \icmlauthor{Kevin Zhu}{algoverse}
\end{icmlauthorlist}

\icmlaffiliation{sjsu}{San Jose State University}
\icmlaffiliation{wpi}{Worcester Polytechnic Institute}
\icmlaffiliation{gmu}{George Mason University}
\icmlaffiliation{algoverse}{Algoverse AI Research}

\icmlcorrespondingauthor{Jerome Francis}{jerome@algoverseairesearch.org}

\icmlkeywords{Latent Chain-of-Thought, Dynamical Systems, Interpretability, Reasoning}

\vskip 0.3in

\icmlkeywords{Machine Learning, ICML, Latent Chain-of-Thought, Interpretability}

\vskip 0.3in




\printAffiliationsAndNotice{
\textsuperscript{1}San Jose State University. 
\textsuperscript{2}Worcester Polytechnic Institute. 
\textsuperscript{3}George Mason University. 
\textsuperscript{4}Algoverse AI Research. \\
\textsuperscript{*}Equal contribution; order decided by random coin flip.
\textsuperscript{$\dagger$}Project Lead. \\
Corresponding author: Jerome Francis \texttt{<jerome@algoverseairesearch.org>}.

$^{1}$ Code Repository:
\url{https://github.com/SabariIyyappan/Latent-CoT-Reasoning-as-Dynamical-Systems}

$^{2}$ Project Page:
\url{https://sabariiyyappan.github.io/Latent-CoT/}
}

\begin{abstract}
Recent latent reasoning methods, such as CODI and COCONUT, face a fundamental interpretability problem: they maintain multiple superimposed candidate traces in the hidden space at each step, unlike explicit-CoT, which follows a single transparent reasoning trace. Existing mechanistic methods show compression, shortcuts, and superposition without explaining how reasoning evolves across latent steps. To address this gap, we model latent token sequences as trajectories in representation space and apply dynamical systems analysis to characterize the evolution of reasoning. Using quantitative measures, such as step-to-step change, direction consistency, and Lyapunov sensitivity, alongside qualitative projections, such as UMAP and DMD/PHATE, we show that latent CoT exhibits structured, non-random dynamics with two distinct stability classes. CODI behaves as a stable attractor, while COCONUT behaves as an unstable expanding system, and SIM-CoT supervision tightens both behaviors without changing the underlying dynamics. This framework advances the interpretability of latent CoT reasoning dynamics and provides actionable insights for improving latent reasoning performance. Code$^{1}$ and Project page$^{2}$ available online.

%

\end{abstract}

\section{Introduction}
\label{introduction}
Latent CoT paradigms such as CODI and COCONUT have consistently outperformed explicit CoT in the performance-compute tradeoff. However, the interpretability of Latent CoT is still an active area of research. Existing mechanistic interpretation methods (logit lens, attention heatmaps, activation patching) reveal relationships between latent tokens and outputs and the role of latent steps through causality, but do not show how reasoning evolves through latent steps.\cite{liang2026latentcot, goyal2025scratchpad}. Prior works demonstrate compression, shortcuts, and superposition in latent CoTs, but do not quantitatively analyze reasoning evolution or highlight quantitative behavior in latent steps.\cite{liang2026latentcot, li2026dynamicslatentcot}.

Dynamical systems provide a principled approach to studying how internal representations evolve during reasoning. Applied to explicit CoT, they help evaluate whether models genuinely reason step-by-step or merely memorize \cite{yu2025stateaware, pham2026rqa}. These works analyze how often the model shifts and visits different states but do not thoroughly measure the rate, direction, or stability of change—factors crucial for understanding reasoning dynamics. The faithfulness of latent CoT—how much latent steps reflect genuine, step-by-step reasoning rather than opaque computation—remains underexplored and lacks rigorous verification. Dynamical systems offer tools to address this gap.
To address this gap, we model latent CoT trajectories as dynamical systems and analyze their stability and representational geometry. Using quantitative metrics and qualitative projections, we investigate whether latent reasoning exhibits structured dynamics and how these dynamics differ across training paradigms.
 
\section{Related Works}
\subsection{Chain-of-Thought and Latent Reasoning}
Chain-of-Thought (CoT) has become the efficient paradigm for LLM reasoning \cite{wei2022cot}, but explicit CoTs are computationally inefficient due to verbose reasoning traces. Latent CoTs were introduced to tackle this problem by skipping the verbose reasoning rationales. Recent latent CoT frameworks such as CODI \cite{shen2025codi} and COCONUT \cite{hao2024coconut} preserve the CoT reasoning in the latent space, achieving better token-efficient performance over explicit CoT.

COCONUT reasons in the latent space by feeding back the last hidden state as next input enabling a breadth-first search over candidate reasoning steps. It is trained using a multi-stage curriculum that progressively replaces explicit CoT tokens with continuous thoughts. CODI instead avoids curriculum learning altogether by jointly training a teacher on explicit CoT and a student on implicit CoT within a single self-distillation stage, transferring reasoning ability via hidden-state at the answer token.

While COCONUT and CODI represent reasoning with latent steps, both methods suffer from the latent instability
problem where we see unstable training as the number of implicit reasoning tokens scale up, leading to latent representations collapsing into homogeneous states that lose semantic diversity. SIM-CoT \cite{wei2025simcot} addresses this instability by adding auxiliary decoder during training to align each implicit latent token with its corresponding explicit reasoning step, enforcing structured semantics in the latent space.

Mechanistic interpretability has emerged as an important direction for understanding the nature of latent CoT and its reasoning behavior, which we discuss in the following section.

\subsection{Mechanistic Interpretability of Latent CoT}
\cite{liang2026latentcot} depicts CODI can retrieve bridge states in a multi-hop reasoning tasks, while relying on compression and short-cut like pathways for longer reasoning chain tasks. \cite{goyal2025scratchpad} shows a ``scratchpad-style'' latent reasoning in CODI where latent states alternate between storage and compute steps. While these works establish that latent reasoning is functionally meaningful, they do not explain the reasoning dynamics.

\cite{li2026dynamicslatentcot} provides a causal analysis of CODI showing that latent reasoning is a two-stream structure where latent states propagate information, while final inputs can bypass computation through direct copy pathways. \cite{liang2026latentcot} has also shown latent-CoTs enable shortcut-like paths to the final answer. These results suggest that latent reasoning contains compression, shortcut routing, and superposition, and they collapse during deeper reasoning tasks, but prior works still lack explanations of ``why'' and ``how'' they emerge in Latent CoTs.

\subsection{Dynamical Systems for Reasoning Dynamics}
Dynamical systems theory provides a mathematical framework for characterizing how states evolve over time, capturing concepts such as attractors, trajectory stability, and divergence. These tools have been widely applied in physics and neuroscience \cite{holmes1990poincare, john2022time} to study complex temporal processes, and have recently been adapted to analyze reasoning behavior in language models.

Viewing reasoning as trajectories in latent spaces has been proven effective for studying reasoning dynamics \cite{yu2025stateaware}. Recent works have shown that explicit CoT reasoning can be analyzed using state-aware and recurrence-based measures to understand how often the model revisits, shifts, or stabilizes in particular reasoning states \cite{yu2025stateaware, pham2026rqa}.

Spectral analysis can be used to represent and cluster reasoning steps semantically to evaluate reasoning dynamics \cite{yu2025stateaware}. Similarly, Pham et al. \cite{pham2026rqa} apply recurrence-based analysis to reasoning traces and show that metrics such as determinism, laminarity, and stalling reveal the reasoning process and the frequency of state visits. These results motivate treating reasoning as a dynamical system rather than as a sequence of independent steps.

However, these approaches have mainly been developed for explicit CoTs, which leaves the study of Latent-CoT reasoning as a dynamical system underexplored. In particular, they do not measure how latent hidden states evolve through representation space in terms of step size, direction consistency, and stability, nor what causes the emergence of phenomena such as compression, shortcuts, and superposition in the Latent-CoT reasoning paradigm. 

Our work addresses this gap by treating latent CoT as a trajectory in latent space and studies its reasoning evolution using dynamical systems.

\section{Methodology}
\begin{figure}[htbp]
    \centering
    \includegraphics[width=0.45\textwidth]{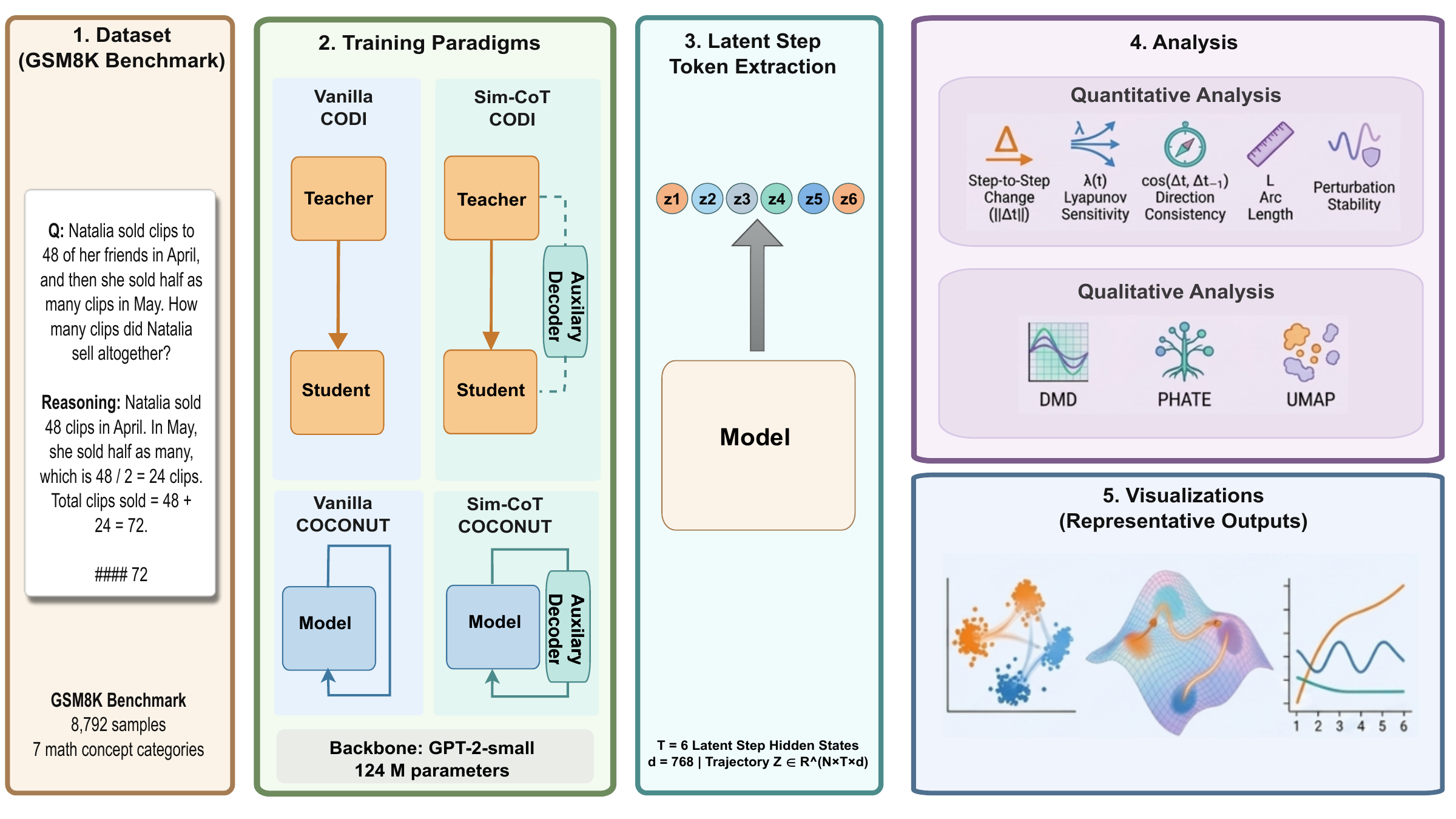}
    \caption{\textbf{Dynamical System Analysis of Latent-CoTs.} (1) GSM8K dataset. (2) Four model configs - CODI (teacher–student self-distillation) and COCONUT (hidden-state feedback loop) in Vanilla and SIM-CoT settings; auxiliary decoder enforces structured semantics. (3) T = 6 latent hidden states form a trajectory in representation space. (4) Quantitative metrics and qualitative projections. (5) Representative latent space and dynamics visualizations.}
    \label{fig:methodology}
\end{figure}

We propose a framework (Figure \ref{fig:methodology}) for analyzing latent chain-of-thought (CoT) reasoning as a dynamical system. The approach extracts intermediate latent representations from reasoning models, constructs trajectories in representation space, and applies dynamical systems analysis to characterize reasoning behaviors.

\subsection{Problem Formulation}
Given an input $x$, a latent CoT model produces a sequence of hidden representations across reasoning steps:
\vspace{-0.5em}
\[ z_1, z_2, \ldots, z_T, \quad z_t \in \mathbb{R}^d, \]
\vspace{-0.5em}
where each $z_t$ denotes the model's internal state at step $t$, and $d$ is the hidden dimension of the model.

This sequence is treated as a trajectory in representation space. 

\subsection{Dimensionality Reduction }
\label{sec:dimred}
The trajectory tensor $Z \in \mathbb{R}^{N \times T \times d}$ lies in a high-dimensional space that makes direct visualization intractable.
Five complementary methods are applied to project latent states into a
low-dimensional space for analysis.
For t-SNE, UMAP and PHATE, the tensor is first flattened to
$[N{\cdot}T,\, d]$ so that all latent states are reduced and then reshaped
 to $[N, T, k]$ to preserve the temporal ordering of each trajectory.

 \textbf{t-SNE} \cite{vandermaaten2008tsne} preserves the local neighborhood structure through pairwise, similarity modeling, making it sensitive to fine-grained clustering among nearby reasoning states.

\textbf{UMAP} \cite{mcinnes2018umap} approximates the underlying data to balance local and global structure, producing stable embeddings suited for continuous trajectories.

Dimensionality reduction is used solely for visualization
(Section~\ref{sec:qualitative_analysis}).
All quantitative metrics are computed directly on the original representations $z_t \in \mathbb{R}^d$.

\subsection{Dynamical Systems Concepts}
\textbf{DMD} \cite{tu2013dmd} treats the latent sequence as a discrete-time dynamical
system and identifies the spatial modes and eigenvalues via an SVD of the snapshot pair
$(X_1, X_2) = (z_{0:T-2},\, z_{1:T-1})$.
The decomposition is computed in a single batched operation across all $N$ trajectories simultaneously. Each trajectory is then projected onto the leading two DMD modes for visualization. The resulting eigenvalues serve as a stability measure in the quantitative analysis (Section~\ref{sec:trajectory_analysis}).

\textbf{PHATE} \cite{moon2019phate} uses a multi-scale diffusion process to capture transitions in high-dimensional data. It uses a diffusion operator based on an affinity graph constructed by kernel methods to calculate the potentials that represent the distances between points, which represents the geometric structure of the underlying manifold. Such a low-dimensional embedding captures both the local structure and global dynamics,
which is also ideal for representing sequential latent paths.

\section{Experimental Configurations}

\subsection{Datasets}

\textbf{GSM8K}
Experiments are conducted on the GSM8K benchmark~\citep{cobbe2021gsm8k},
a collection of grade-school math word problems requiring multi-step
arithmetic reasoning.
The full GSM8K dataset (train and test splits combined, 8,792 samples) is used across seven math concept categories: Geometry (210),
Rates \& Speed (675), Percentages \& Ratios (1,266), Money \& Pricing
(2,741), Fractions \& Decimals (1,045), Multiplication \& Division (224), and Arithmetic \& Multi-step (2,631).
Questions are assigned to categories via priority-ordered keyword matching,
with more specific categories checked before broader ones to prevent improper labeling.
Five hundred stratified samples are taken per category. Categories with
fewer available examples contribute all remaining samples.
Ground-truth answers are parsed from the \texttt{"\#\#\#\#"} delimiter in GSM8K annotations.

\subsection{Models}

CODI and COCONUT are both GPT-2-small backbones (124 M parameters, 12 layers, 12 heads, hidden size 768, vocabulary extended to 50 260 with three latent special tokens\texttt{<|start-latent|>}, \texttt{<|latent|>}, and\texttt{<|end-latent|>}) trained on GSM8K~\citep{hao2024coconut,shen2025codi}. Vanilla checkpoints are taken from \texttt{ModalityDance/latent-tts-coconut}\footnote{\url{https://huggingface.co/ModalityDance/latent-tts-coconut}} and \texttt{ModalityDance/latent-tts-codi}\footnote{\url{https://huggingface.co/ModalityDance/latent-tts-codi}}; CODI additionally loads a self-distillation projection module (\texttt{prj.pt}) at inference time that maps each latent hidden state back into the input-embedding space. SIM-CoT variants of both methods are taken from \texttt{internlm/SIM\_COT-GPT2-Coconut}\footnote{\url{https://huggingface.co/internlm/SIM_COT-GPT2-Coconut}} and \texttt{internlm/SIM\_COT-GPT2-CODI}\footnote{\url{https://huggingface.co/internlm/SIM_COT-GPT2-CODI}}~\citep{wei2025simcot}. Each model generates a fixed sequence of $T = 6$ latent reasoning steps per input. Text generation uses greedy decoding with a maximum of 512 output tokens. All experiments use a fixed random seed of 42.

\subsection{Analysis of Latent CoT trajectories}
Analysis is performed at 2 levels: qualitative inspection
of geometric structure and quantitative characterization of
dynamics.

\subsubsection{Qualitative Analysis}
\label{sec:qualitative_analysis}

Reduced representations from Section~\ref{sec:dimred}
are visualized as sequences of points in $\mathbb{R}^2$ and $\mathbb{R}^3$
with temporal ordering preserved.
These plots expose geometric phenomena including smooth
progression through state space, directional pivots (using centroids of latent representations), convergence, and separation between correct and incorrect trajectories.

\subsubsection{Quantitative Analysis}
\label{sec:trajectory_analysis}

Trajectory dynamics are characterized by two groups of
metrics: \emph{step-based metrics} that measure how the
trajectory moves between consecutive steps, and
\emph{stability metrics} that characterize sensitivity and
attraction to fixed points.

\textbf{Step-based Metrics}
\\
\textbf{Step-to-step change} measures the magnitude of
displacement at each transition:
\begin{equation}
    \|\Delta_t\| = \|z_{t+1} - z_t\|_2.
\end{equation}
Large values indicate substantial representational transitions, while smaller values indicate reduced movement between consecutive latent states. Although decreasing displacement is consistent with convergence toward a stable region of representation space, it does not necessarily imply convergence toward a correct solution.

\textbf{Direction consistency} measures whether consecutive
displacements point in the same direction:
\begin{equation}
    C_t = \cos(\Delta_t,\, \Delta_{t-1}).
\end{equation}
Values near $+1$ indicate aligned transitions, values near $0$ indicate directional pivots, and values near $-1$ indicate reversals. We interpret this metric as describing trajectory geometry rather than reasoning quality, since effective reasoning may involve branching, revision, or exploration of multiple solution paths.

\textbf{Arc length} summarizes total path complexity as the
cumulative displacement over all transitions:
\begingroup
\setlength{\abovedisplayskip}{2pt}
\setlength{\belowdisplayskip}{2pt}
\[
   L = \sum_{t=1}^{T-1} \|z_{t+1} - z_t\|_2.
\]
\endgroup
Arc length summarizes the cumulative displacement traversed by a latent trajectory and serves as a measure of overall trajectory complexity.
\\\\
\textbf{Stability Metrics}
\\
\textbf{Lyapunov Sensitivity (Surrogate)}
To estimate local stability without re-running the model, a trajectory-based surrogate is computed from the ratio of onsecutive step magnitudes:
\begin{equation}
    \lambda(t) = \log \frac{\|z_{t+1} - z_t\|}{\|z_t - z_{t-1}\|},
    \label{eq:lyapunov}
\end{equation}
where $\|\cdot\|$ denotes the $\ell_2$ norm.
A positive $\lambda(t)$ indicates the trajectory is locally diverging.
A negative $\lambda(t)$ indicates convergence.
$\lambda(t) = 0$ indicates neutral stability, where the step magnitude remains constant and the trajectory is neither converging nor diverging.
Positive values indicate local expansion in trajectory magnitude, while negative values indicate contraction. We interpret these behaviors as potential signatures of exploratory and stabilizing dynamics respectively, although additional causal validation would be required to establish a direct correspondence with reasoning processes.


\textbf{Perturbation stability} re-runs inference with
Gaussian noise injected into input embeddings and measures divergence from the clean trajectory at each step. Growing divergence indicates a sensitivity to initial conditions, demonstrating the robustness of latent CoT tokens during reasoning.

\section{Results}
\label{sec:results}

Results are reported for COCONUT and CODI methods for both Vanilla COT and SIM-CoT training paradigms. The analysis is split into two parts: (1) \textit{Qualitative dynamics}, where we visualize latent states using DMD, PHATE, and UMAP to understand how each model moves through latent space across reasoning steps; and (2) \textit{Quantitative dynamics}, where we measure step-to-step change, direction consistency, and fixed-point distances to compare how the two models behave differently.

* Concept-wise plots (Arc Length, Perturbation Stability, Step-to-Step Change, Direction Consistency, Lyapunov Sensitivity) for GSM8K are provided in
Appendix~\ref{app:gsm8k_plots}.

\subsection{Quantitative Results}
\label{sec:metrics_res}

\begin{figure}[ht]
\centering
\renewcommand{\arraystretch}{1}
\begin{tabular}{c c c}
 & \textbf{CODI} & \textbf{COCONUT} \\[0.5em]
\raisebox{0.5cm}{\rotatebox{90}{\textbf{Vanilla}}} &
\includegraphics[width=0.30\linewidth,trim={0 0 0 1.2cm},clip]{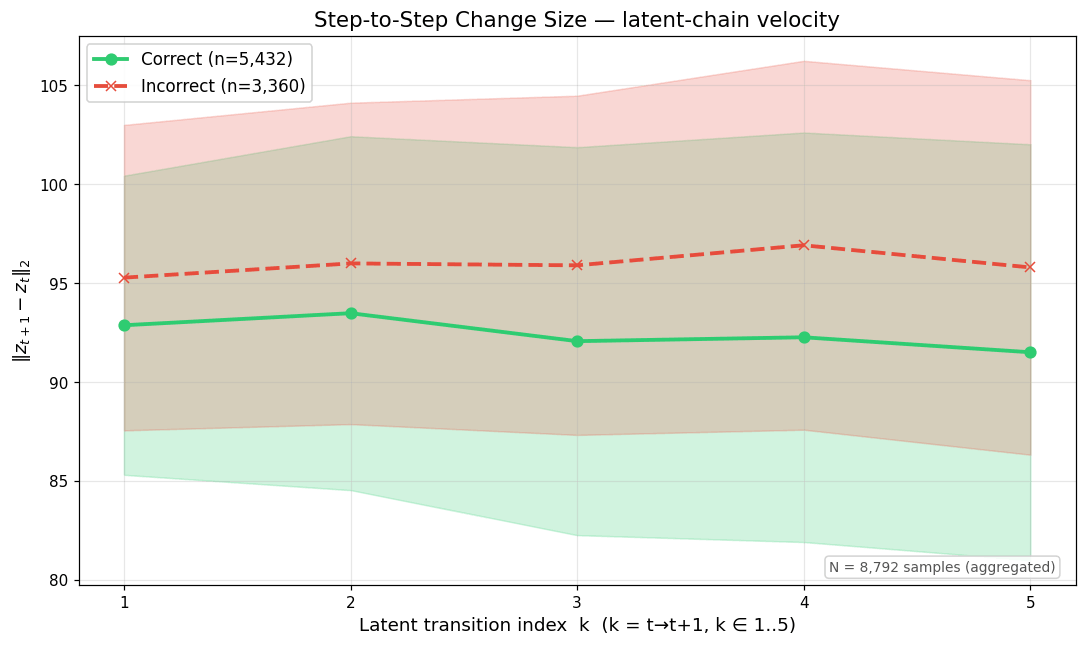}
&
\includegraphics[width=0.30\linewidth,trim={0 0 0 1.2cm},clip]{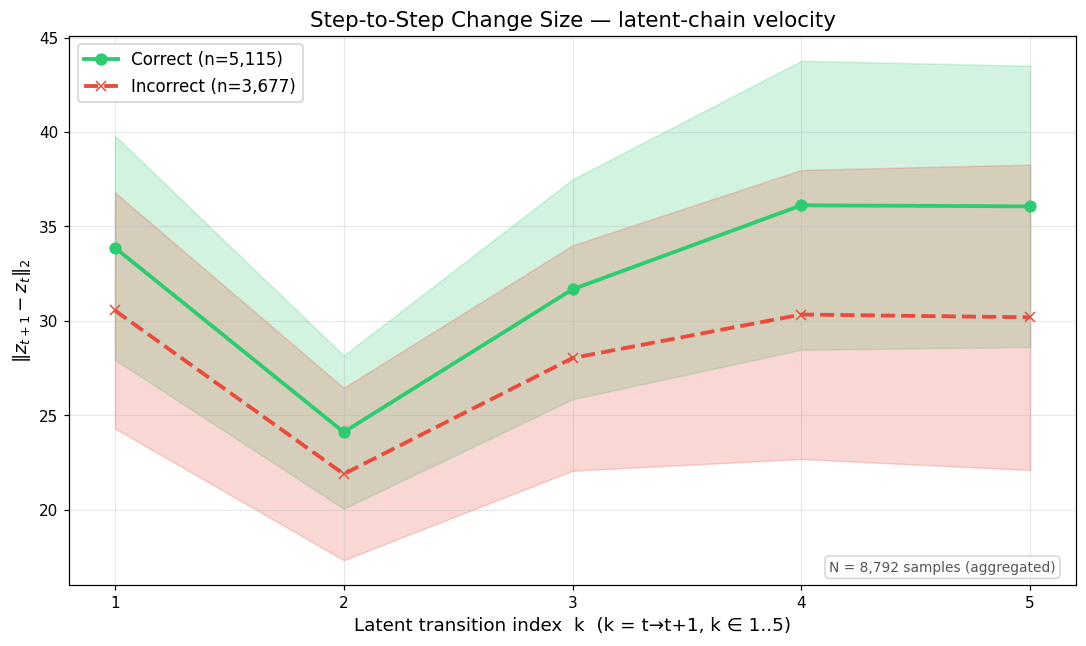}
\\[0.5em]
\raisebox{0.5cm}{\rotatebox{90}{\textbf{SIM-CoT}}} &
\includegraphics[width=0.30\linewidth,trim={0 0 0 3cm},clip]{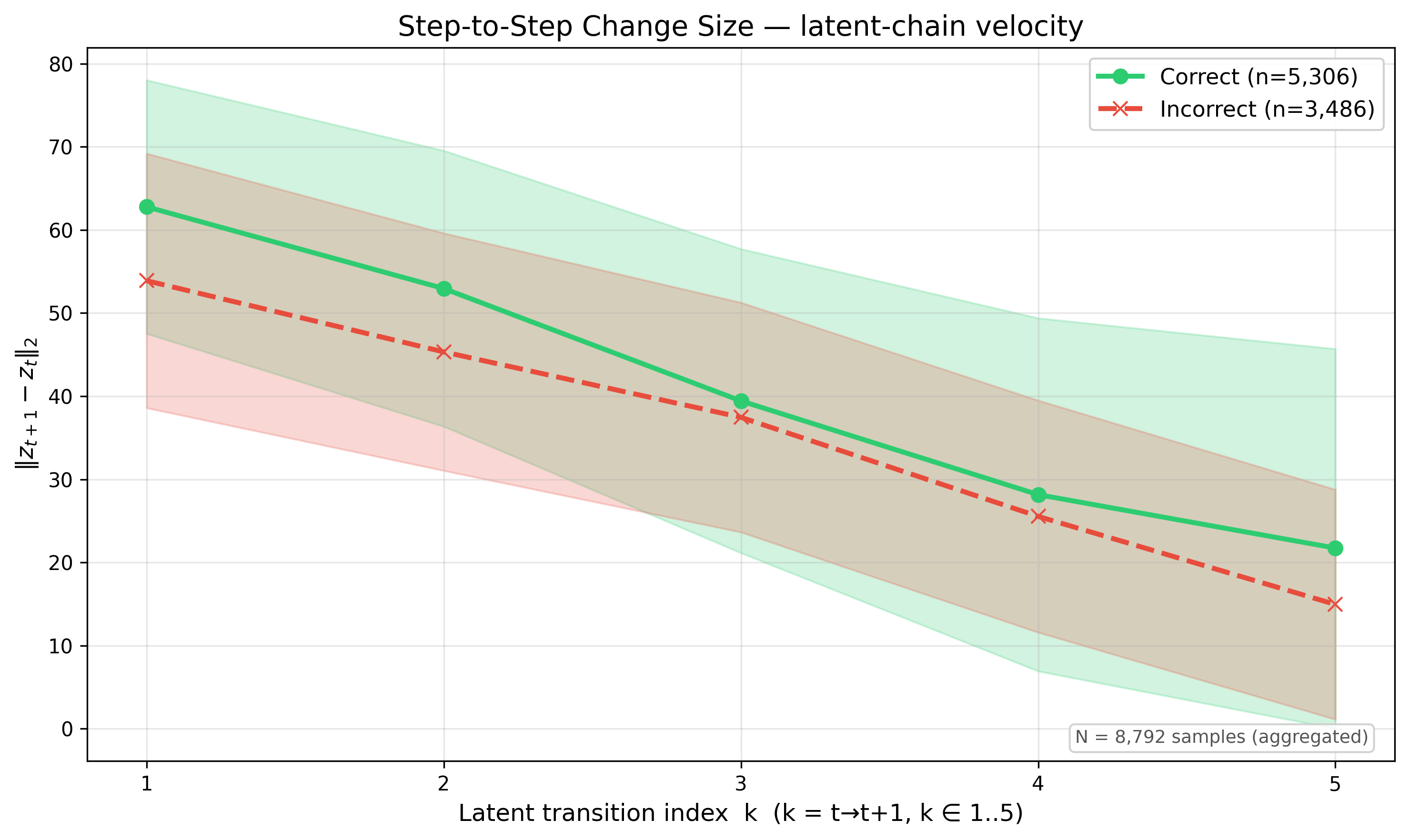}
&
\includegraphics[width=0.30\linewidth,trim={0 0 0 3cm},clip]{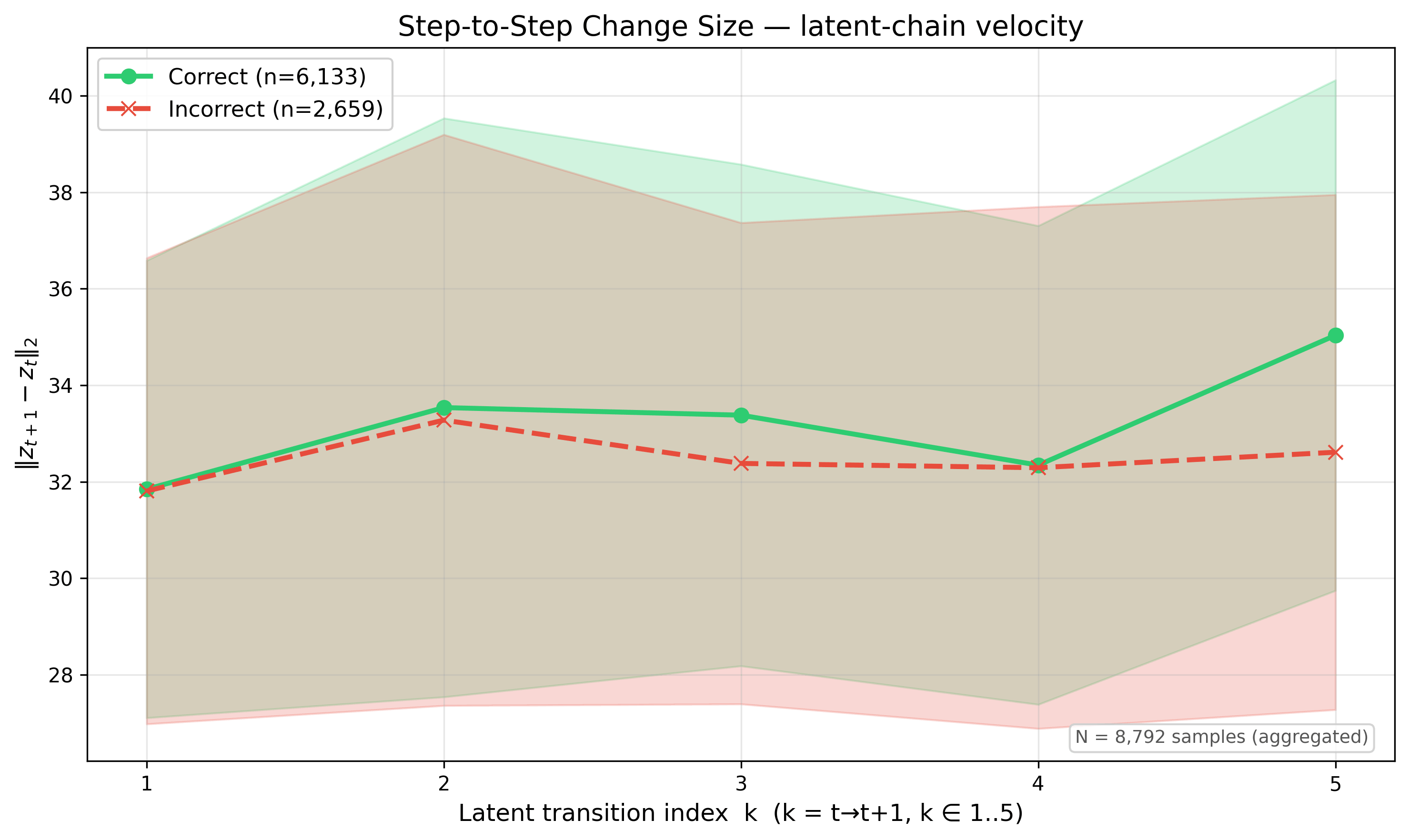}
\end{tabular}
\caption{
\textbf{Step-to-Step change for methods (CODI, COCONUT) under Vanilla + SIM-CoT settings}: Overall, the CODI method exhibits neutral or convergent behavior , while COCONUT shows premature convergence in the vanilla setting. Both the methods have more stable curves in SIM-CoT setting compared to their Vanilla counterparts. 
}
\label{fig:step2stepchange_2x2}
\end{figure}

\paragraph{Step-to-Step Change}
The analysis here (as seen in Figure \ref{fig:step2stepchange_2x2}) measures the consecutive differences between latent reasoning steps. 
For the CODI method, we see an uniform curve with variance for vanilla CoT setting. In contrast, SIM-CoT setting displays a monotonically decreasing curve. This decreasing characteristic demonstrates that latent CoT tokens start to become similar, a potential sign for convergence in the end phase (solution). For the COCONUT method, the vanilla CoT setting shows a drop in the transition from $2^{nd}$ to $3^{rd}$ latent token, a possible premature convergence in the reasoning process. Whereas, we see a stable behavior in SIM-CoT setting.

\begin{figure}[ht]
\centering
\renewcommand{\arraystretch}{1}
\begin{tabular}{c c c}
 & \textbf{CODI} & \textbf{COCONUT} \\[0.5em]
\raisebox{0.5cm}{\rotatebox{90}{\textbf{Vanilla}}} &
\includegraphics[width=0.30\linewidth,trim={0 0 0 1.93cm},clip] {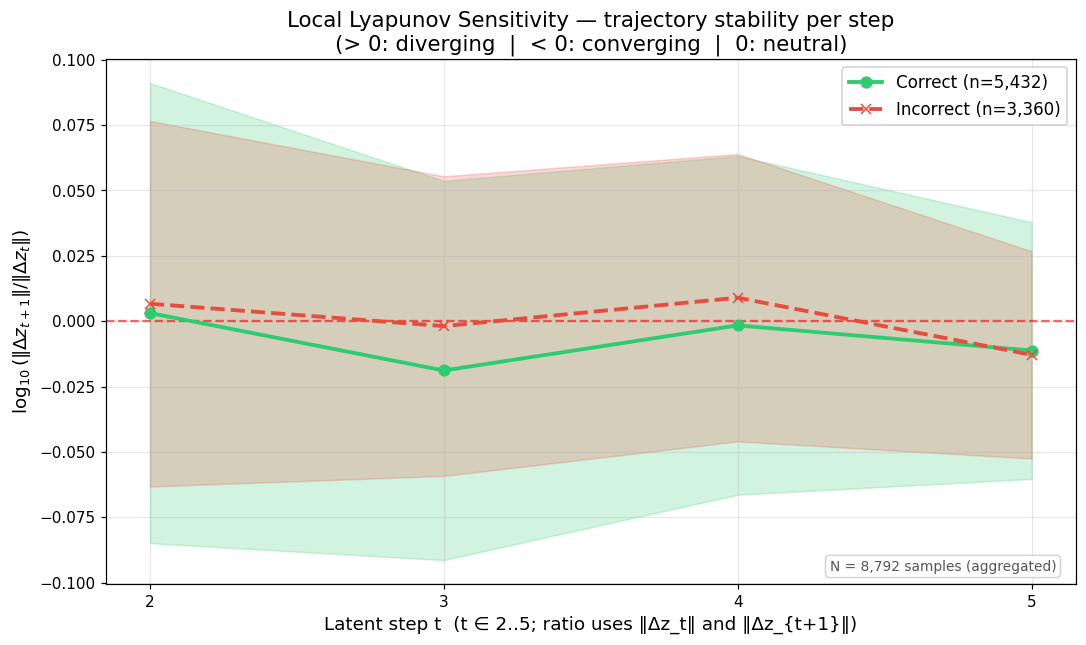}
&
\includegraphics[width=0.30\linewidth,trim={0 0 0 1.93cm},clip]{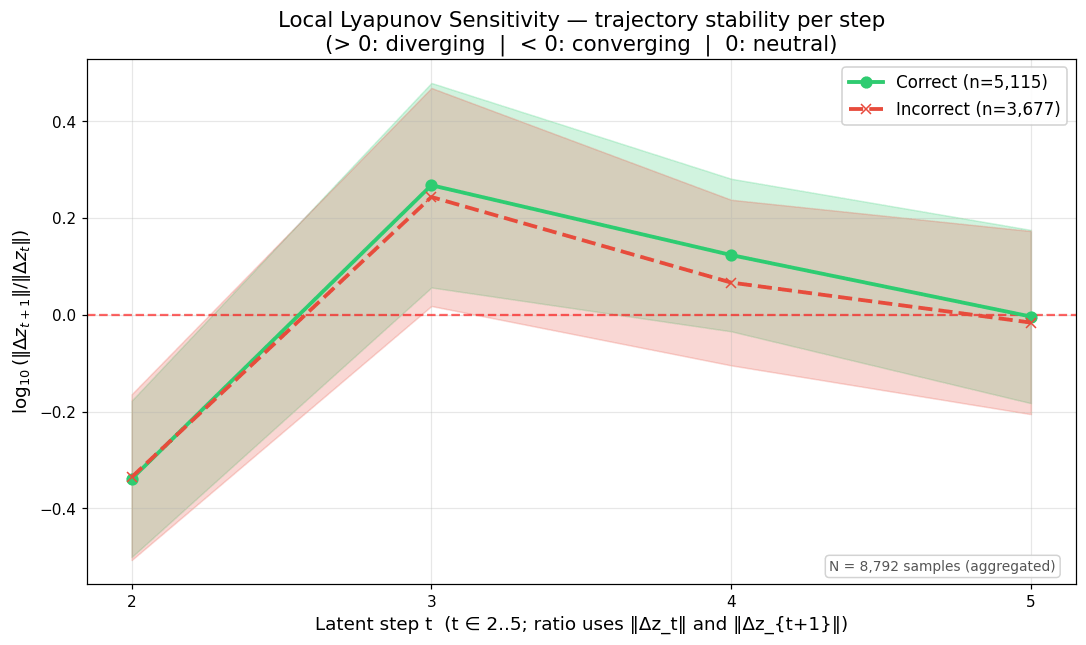}
\\[0.5em]
\raisebox{0.5cm}{\rotatebox{90}{\textbf{SIM-CoT}}} &
\includegraphics[width=0.30\linewidth,trim={0 0 0 5.7cm},clip]{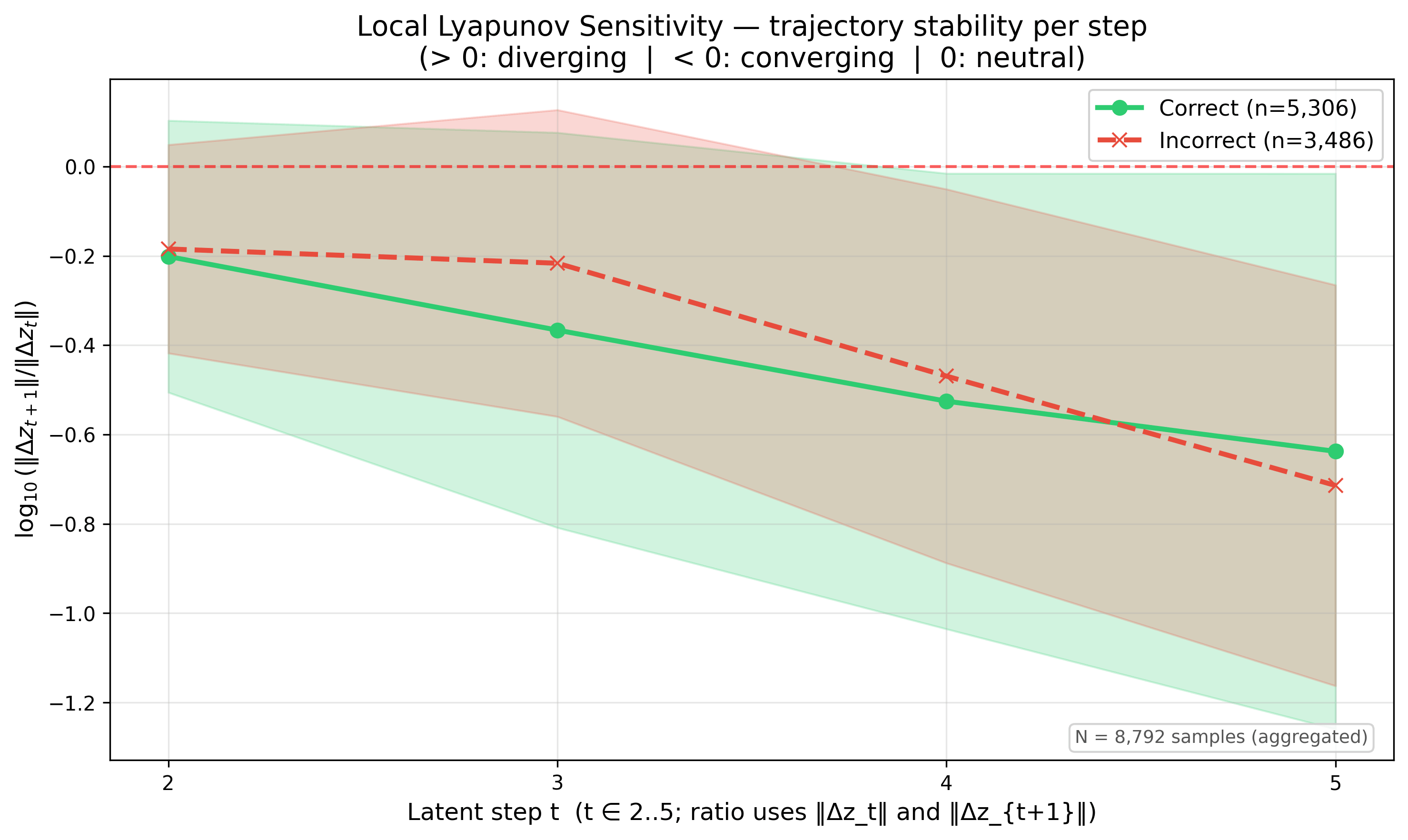}
&
\includegraphics[width=0.30\linewidth,trim={0 0 0 5.7cm},clip]{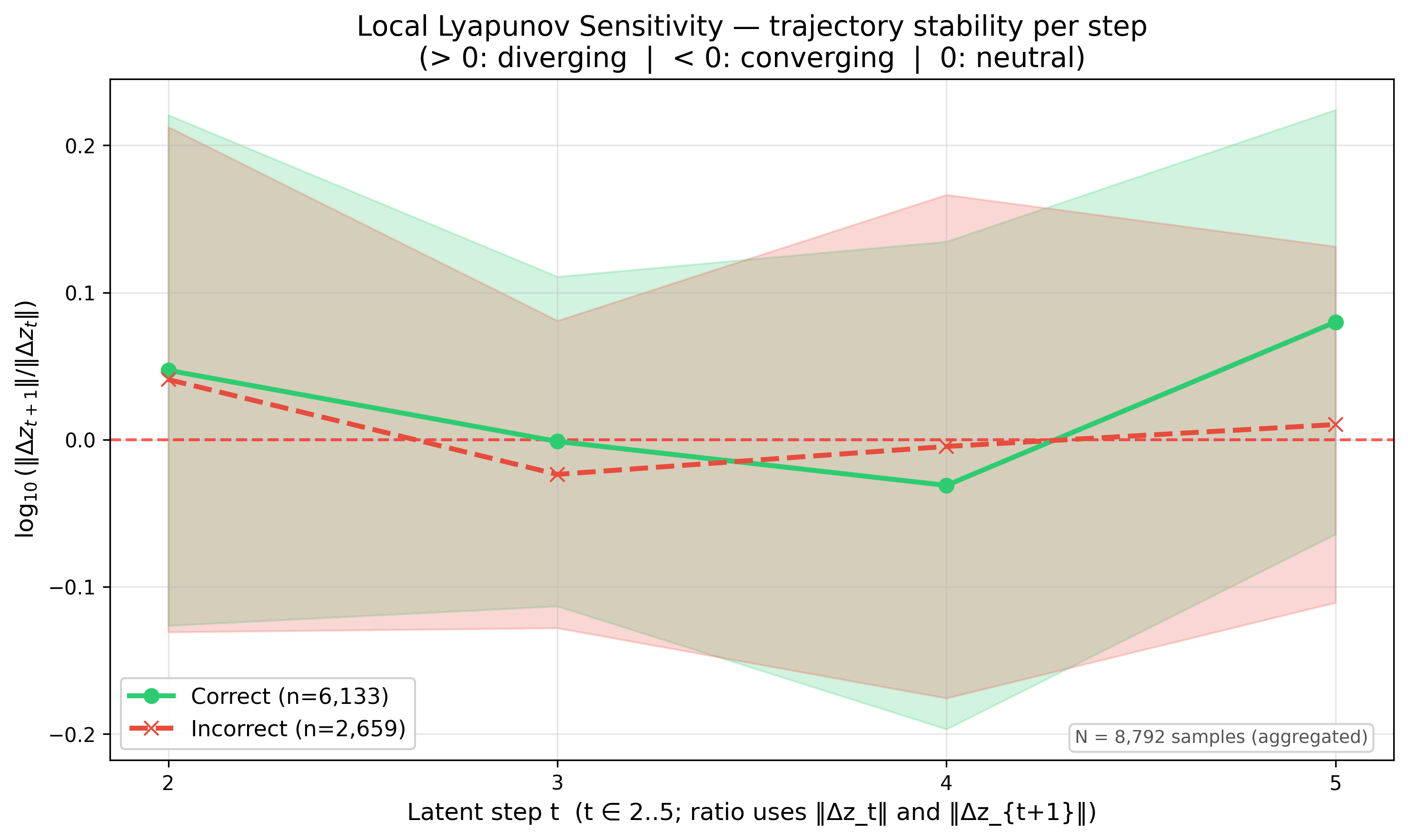}
\end{tabular}
\caption{
\textbf{Lyapunov Sensitivity for methods (CODI, COCONUT) under Vanilla + SIM-CoT settings}: CODI exhibits stable, near-neutral or converging behavior across both settings, while COCONUT shows a sharp divergence (local instability) at t=3 in the Vanilla setting. SIM-CoT supervision suppresses this instability in COCONUT and deepens the convergence in CODI. \\(> 0: diverging | < 0: converging | 0: neutral )
}
\label{fig:lyapunovsensitivity_2x2}
\end{figure}

\paragraph{Lyapunov Sensitivity}
Lyapunov sensitivity (as seen in Figure \ref{fig:lyapunovsensitivity_2x2}) measures the local stability of the latent trajectory. Positive values indicate the trajectory is locally diverging (the model is exploring multiple candidate reasoning paths at that step), while negative values indicate it is locally contracting, i.e. committing to a single path.
\\
For the CODI method, we see a monotonically decreasing curve with negative values across all steps in the Vanilla CoT setting. In the SIM-CoT setting, the curve descends deeper into the negative range, indicating that the supervision strengthens this convergence. For the COCONUT method, the Vanilla CoT setting exhibits a sharp positive spike at $t=3$, a localized expansion that captures the model branching across candidate reasoning paths at the mid-chain transition before settling, a genuine exploration phase native to COCONUT's curriculum-trained dynamics. Whereas, in the SIM-CoT setting, this spike disappears and the curve flattens, showing that the supervision replaces the exploratory mid-chain step with a deterministic transition by anchoring each latent token to its corresponding textual reasoning step.
\\

\FloatBarrier

\paragraph{Direction Consistency}
Direction consistency measures the cosine between consecutive latent transitions,
$C_t = \cos(\Delta_t, \Delta_{t-1})$. Values near $+1$ indicate smooth forward motion, values near $0$ indicate orthogonal pivots, and values near $-1$ indicate the trajectory reverses direction at every step. For CODI, the Vanilla setting exhibits predominantly negative values, whereas SIM-CoT shifts the trajectory closer to orthogonal transitions. For COCONUT, the Vanilla setting transitions from opposing to near-orthogonal directions before returning toward opposing transitions. SIM-CoT produces a more consistent directional profile across latent steps.


\begin{figure}[ht]
\centering
\renewcommand{\arraystretch}{1.0} 
\begin{tabular}{c c c}
 & \textbf{CODI} & \textbf{COCONUT} \\ 

\raisebox{0.25cm}{\rotatebox{90}{\textbf{Vanilla}}} &
\includegraphics[width=0.30\linewidth,trim={0 0 0 1.93cm},clip]{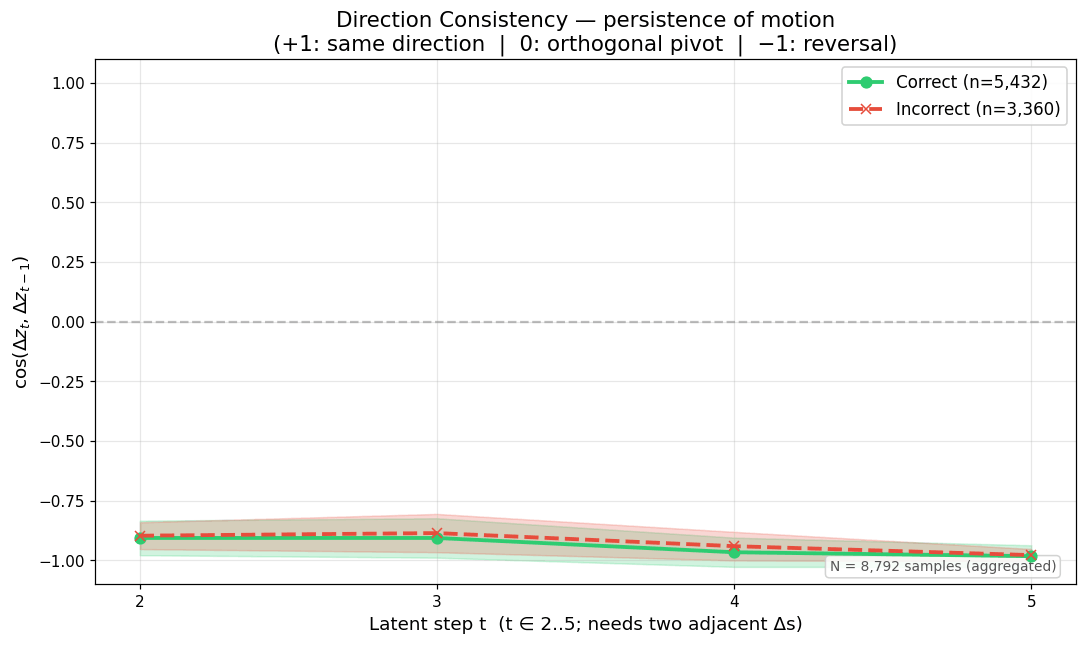} &
\includegraphics[width=0.30\linewidth,trim={0 0 0 1.93cm},clip]{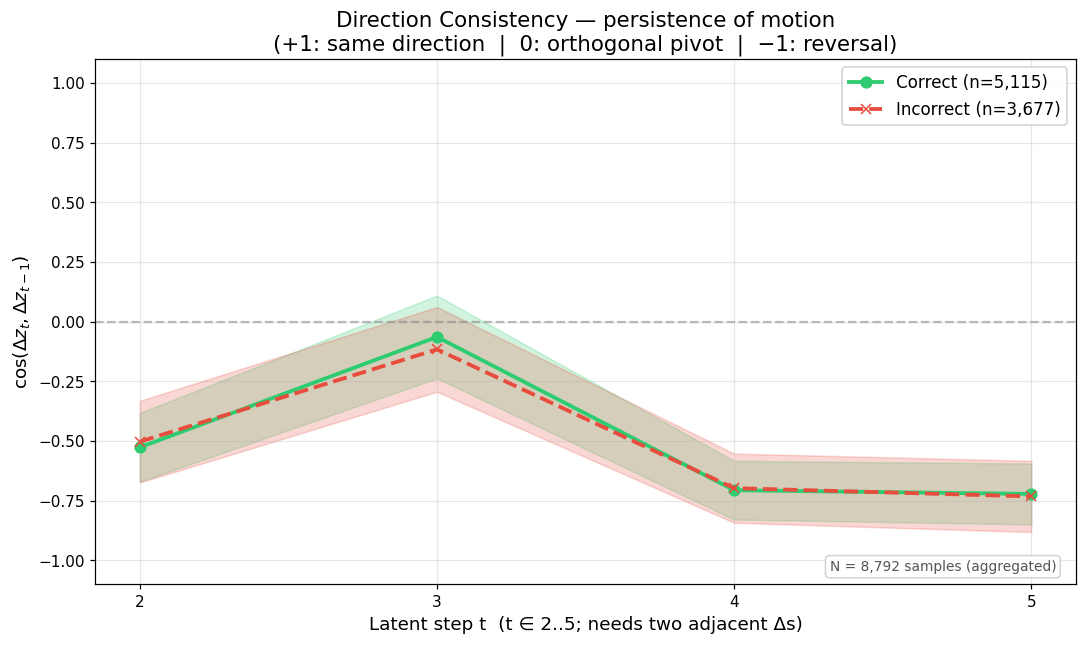} \\ 

\raisebox{0.25cm}{\rotatebox{90}{\textbf{SIM-CoT}}} &
\includegraphics[width=0.30\linewidth,trim={0 0 0 5.5cm},clip]{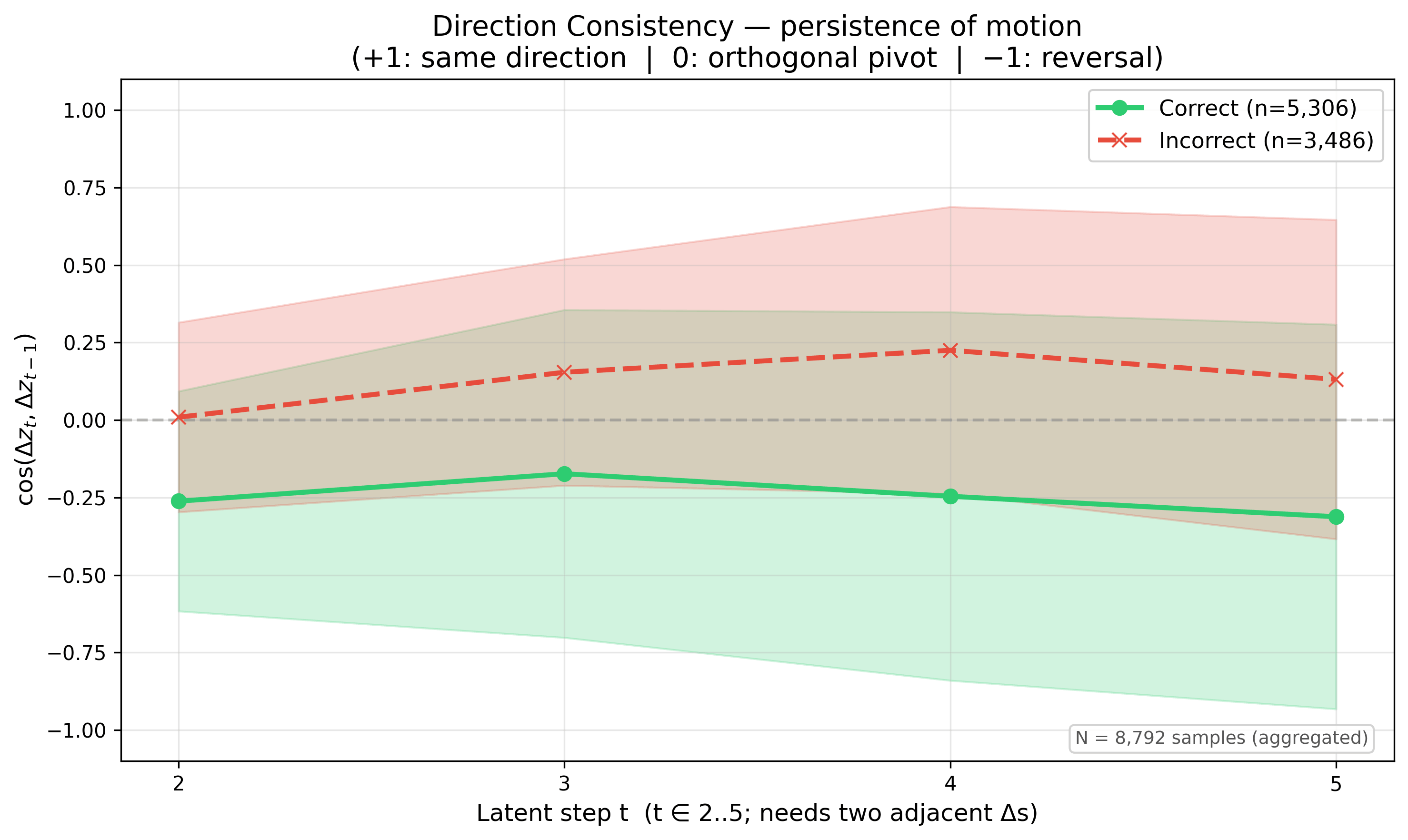} &
\includegraphics[width=0.30\linewidth,trim={0 0 0 5.5cm},clip]{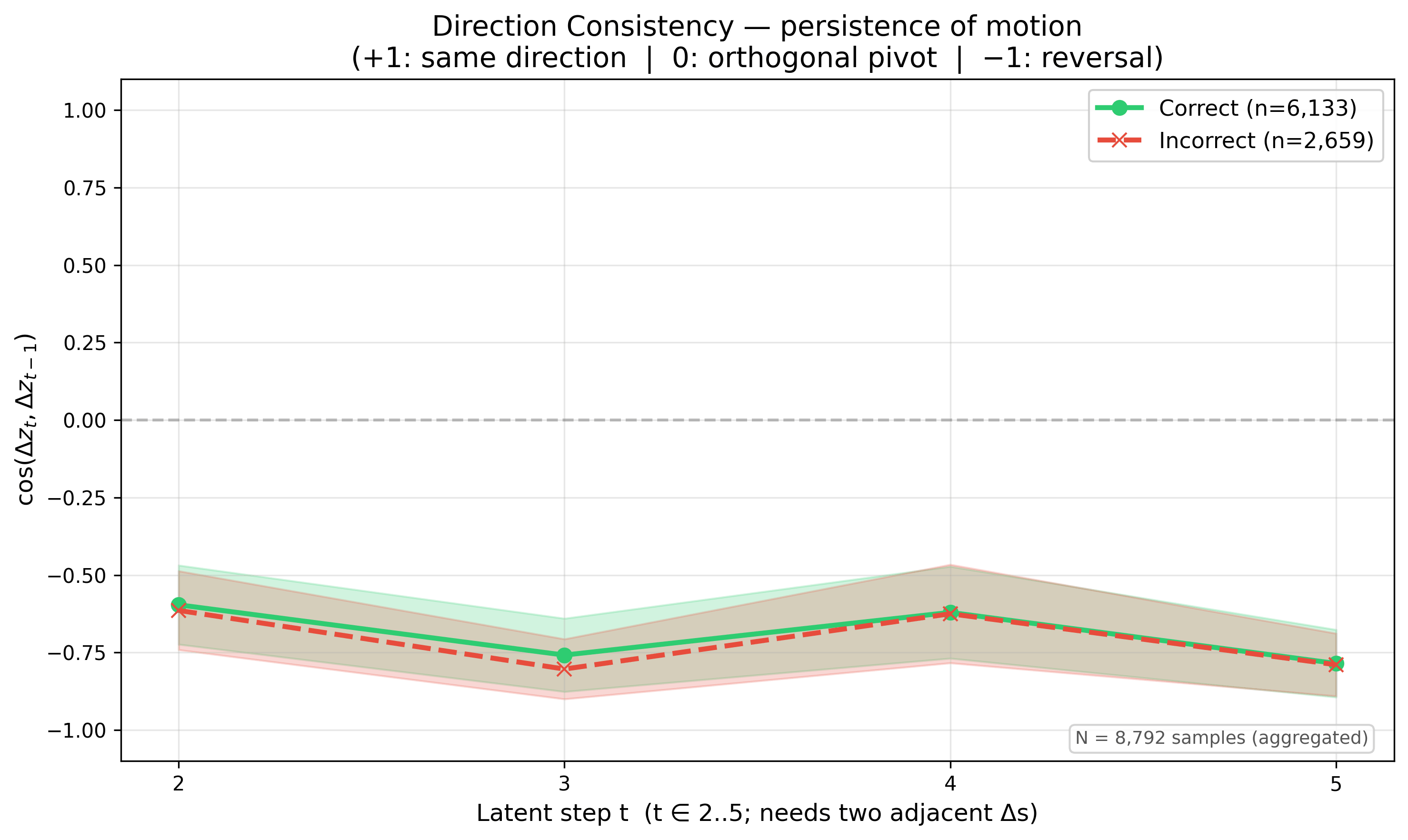}
\end{tabular}
\vspace{-0.5em}
\caption{
\textbf{Direction consistency $C_t = \cos(\Delta_t, \Delta_{t-1})$ across latent steps for methods (CODI and COCONUT) under Vanilla + SIM-CoT settings}: In the Vanilla setting, CODI shows consistently opposing transitions, while COCONUT transitions become near orthogonal at t=3 before reversing again. Under SIM-CoT, CODI shifts to near orthogonal transitions and COCONUT maintains consistent opposition.(+1: Same direction | 0: Orthogonal pivot | -1: Reversal)
}
\label{fig:directionconsistency_2x2}
\end{figure}

\textbf{Arc length and Perturbation stability} Arc length summarizes the total displacement traversed by the latent trajectory, reflecting overall reasoning effort.
Perturbation stability measures trajectory divergence
$\lVert z_t^{\text{perturbed}} - z_t^{\text{clean}}\rVert_2$ under Gaussian noise injected into the input embeddings, tested at $\sigma \in \{0.01, 0.1, 1.0\}$.Results and additional discussions are added in the Appendix ~\ref{sec:A.3.1_arc_len} and ~\ref{sec:A.3.2_perturb}.

\subsection{Qualitative Analysis}
\label{sec:traj_geo}

\begin{figure}[ht]
\centering
\begin{minipage}{0.30\linewidth}
\centering
\includegraphics[width=\linewidth,trim={0 0 0 1cm},clip]{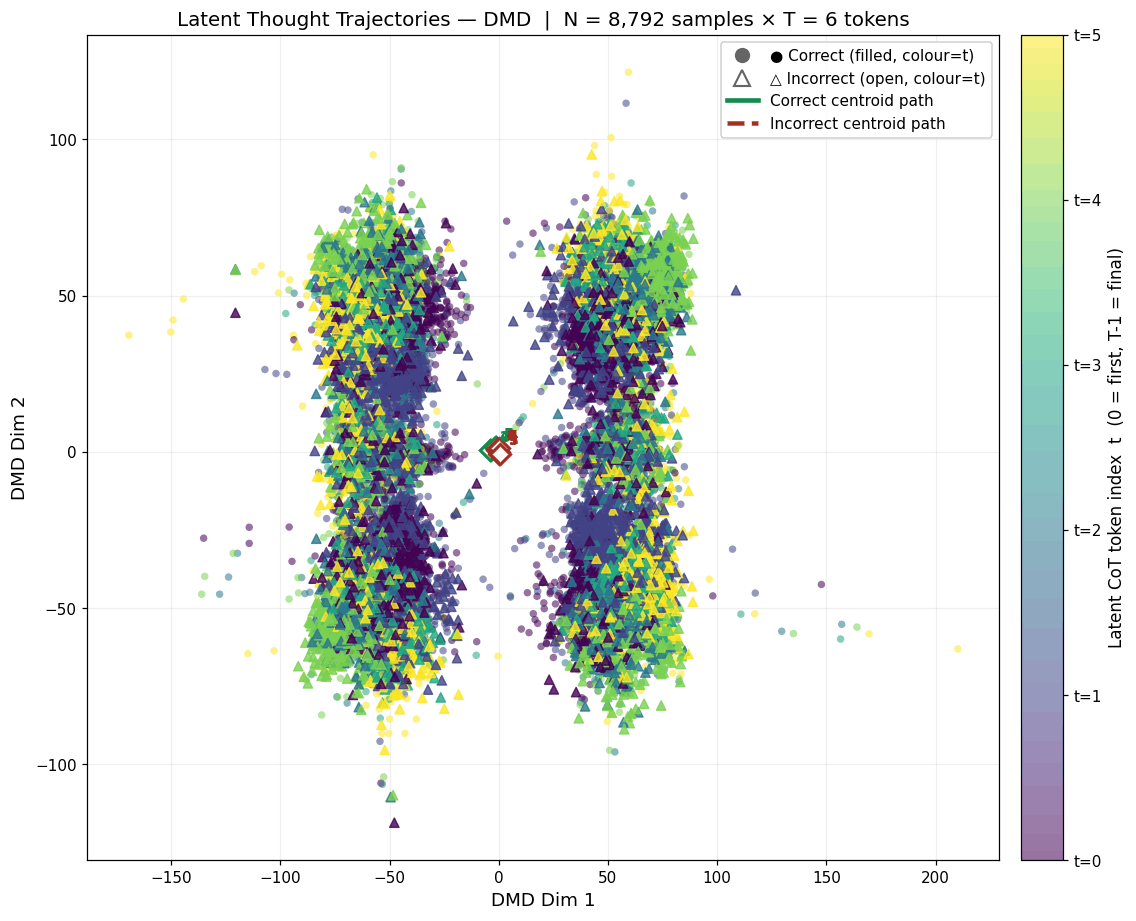}
\end{minipage}
\hspace{0.02\linewidth}
\begin{minipage}{0.30\linewidth}
\centering
\includegraphics[width=\linewidth,trim={0 0 0 1cm},clip]{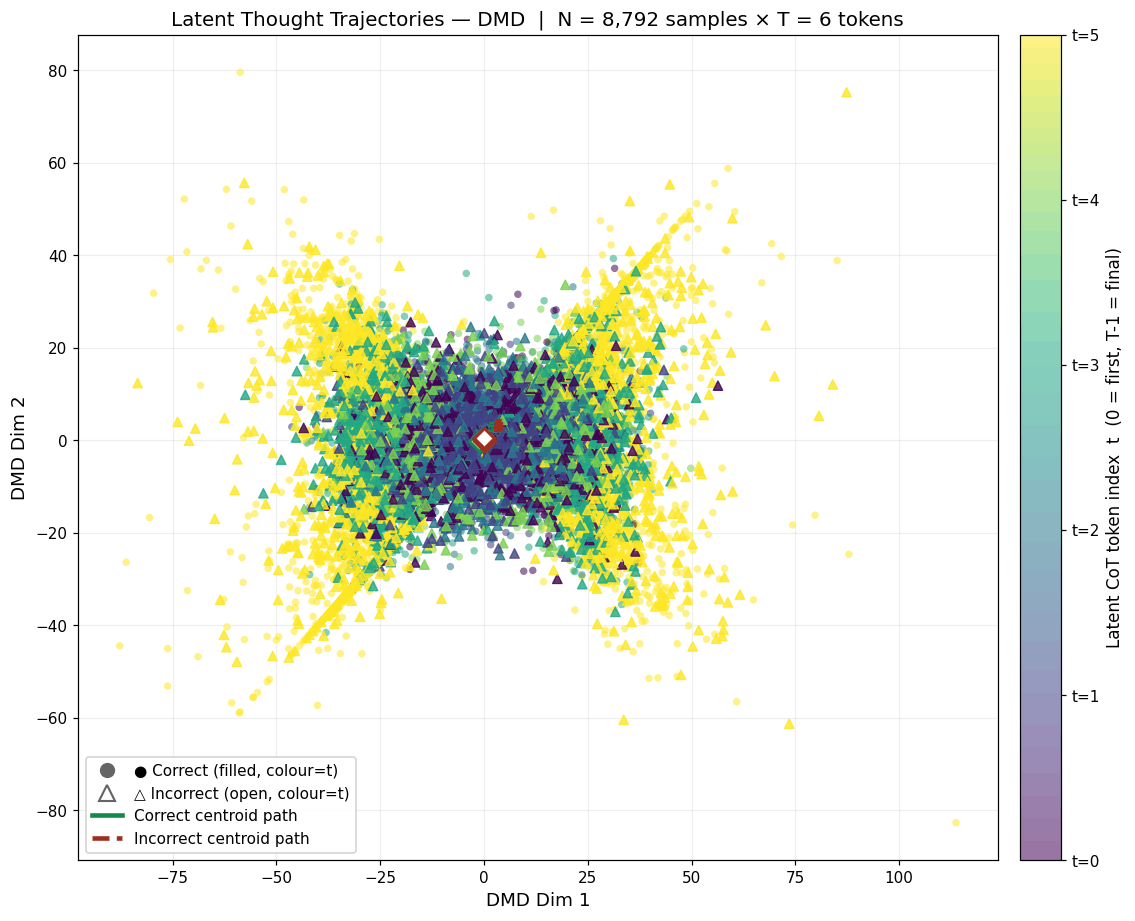}

\end{minipage}
\caption{\textbf{DMD trajectory projections across latent steps for methods (CODI (left) and COCONUT(right))}: COCONUT exhibits a butterfly pattern with latent states diverging outward across steps, reflecting an unstable system exploring multiple reasoning paths simultaneously. CODI shows tightly packed representations clustering into two dense, well-separated regions across all latent steps, indicating a convergent and stable reasoning dynamic.\\(N = 8,792 samples x T = 6 tokens)
}
\label{fig:traj_dmd_gsm8k}
\end{figure}

\paragraph{DMD} (as seen in Figure~\ref{fig:traj_dmd_gsm8k}) projects each latent trajectory onto its dominant modes of variation, showing how the latent representations are spatially organized across the reasoning chain. Tightly clustered latent states indicate contraction toward a bounded region, while spreading latent states indicate expansion away from it.
\\
For the CODI method, the vanilla CoT setting shows a two-lobe pattern with the latent CoT tokens tightly packed into both lobes across all steps, a sign of stable attractor-like organization where the trajectory remains bounded throughout the chain. In the SIM-CoT setting, the two-lobe pattern is preserved with even tighter clustering, indicating that the supervision reinforces this bounded
behavior.
\\
For the COCONUT method, the vanilla CoT setting exhibits a butterfly pattern where the latent states start near the center at $t=0$ and spread outward by $t=5$, a sign of expanding behavior where the model explores across latent space as the chain progresses. Whereas, in the SIM-CoT setting, the butterfly geometry is retained but with reduced spread at later steps, showing that the supervision constrains the exploration without changing the underlying geometric structure.

\begin{figure}[ht]
\centering
\begin{minipage}{0.30\linewidth}
\centering
\includegraphics[width=\linewidth,trim={0 0 0 0.8cm},clip]{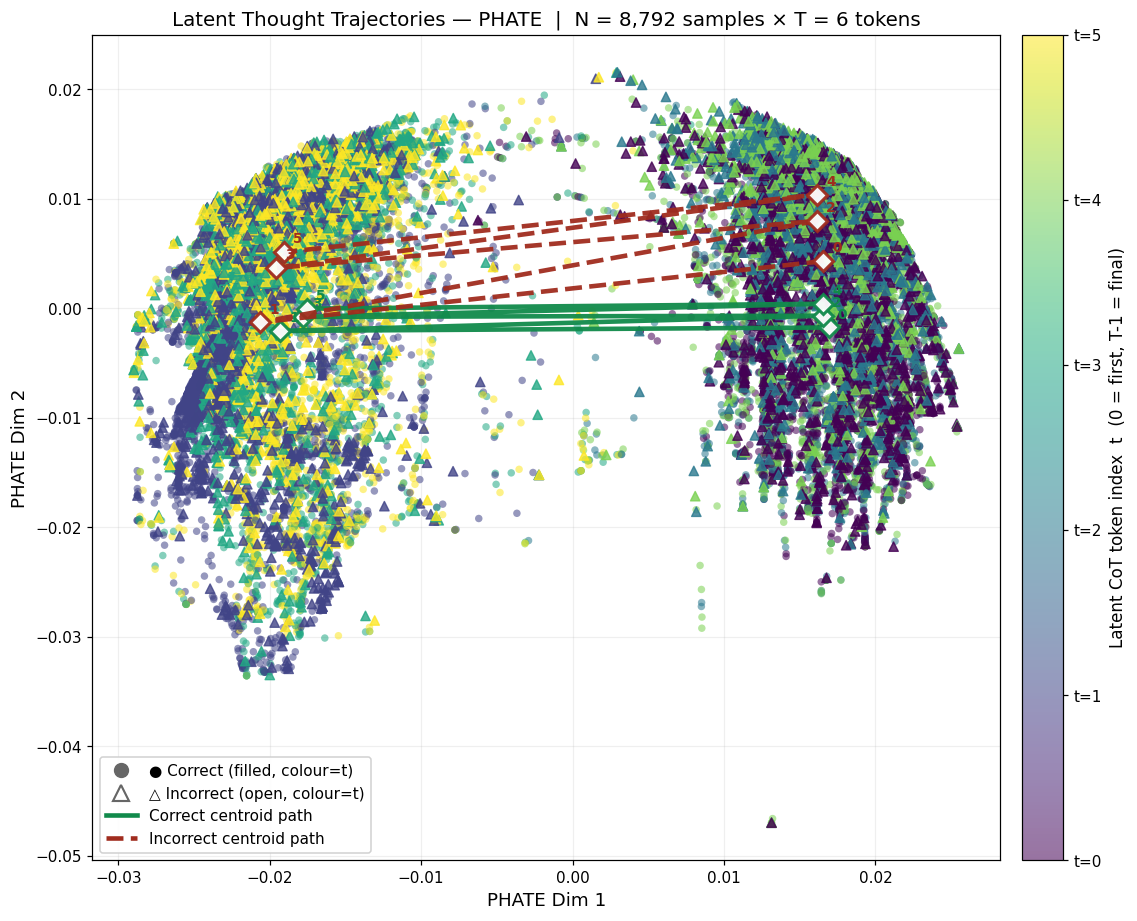}
\end{minipage}
\hspace{0.02\linewidth}
\begin{minipage}{0.30\linewidth}
\centering
\includegraphics[width=\linewidth,trim={0 0 0 0.8cm},clip]{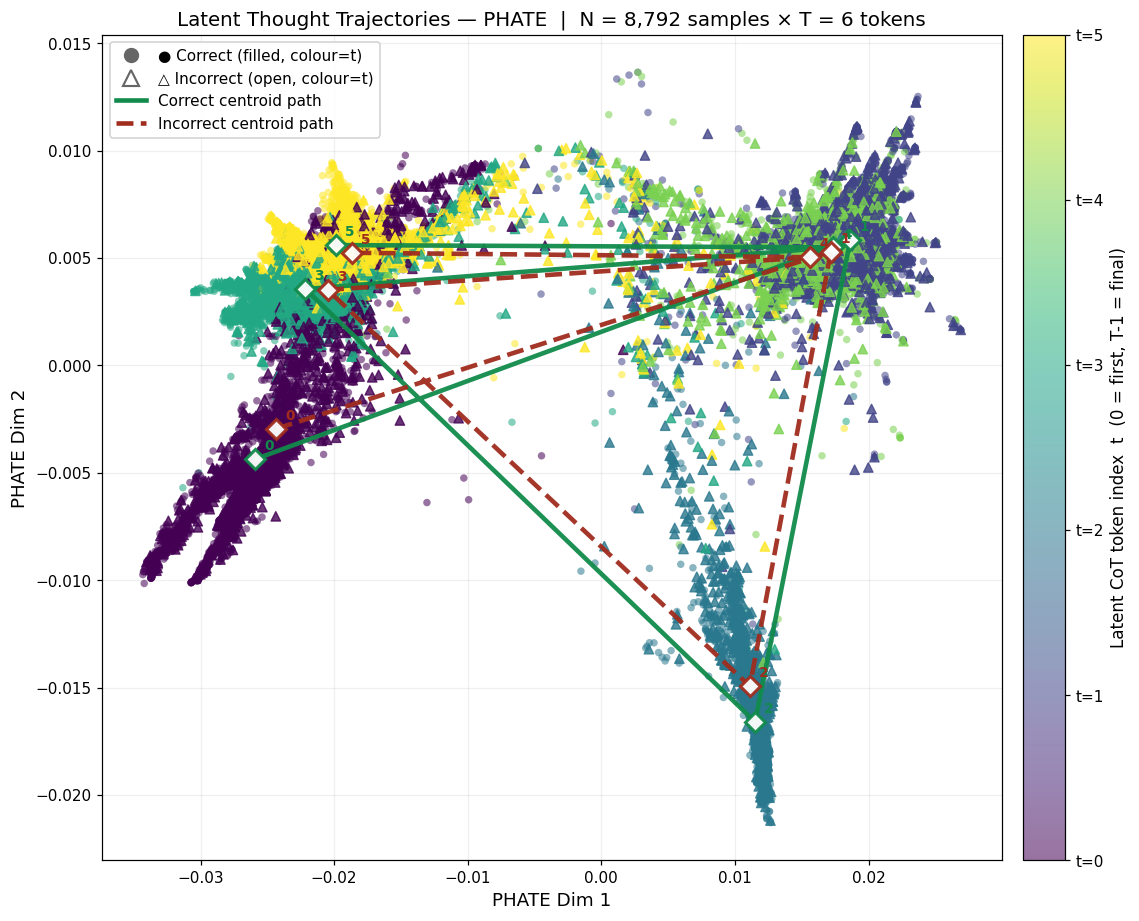}
\end{minipage}
\caption{\textbf{PHATE trajectory projections across latent steps for methods (CODI (left) and COCONUT(right))}: COCONUT exhibits each latent step occupying distinct regions of the manifold, with beginning and end tokens positioned in closer proximity-suggesting the formation of shortcut pathways in the reasoning process. CODI shows interleaved representations across steps, consistent with its convergent behavior observed in the DMD projection. (Number of samples = 8,792 samples, 6 tokens)
}
\label{fig:traj_phate_gsm8k}
\end{figure}

\paragraph{PHATE} (as seen in Figure~\ref{fig:traj_phate_gsm8k}, right) embeds the latent trajectory using a multi-scale diffusion process that preserves both local neighborhoods and the global manifold structure, making it suitable for showing how latent representations move through different regions of the reasoning manifold.
\\
For the CODI method, the vanilla CoT setting shows a two-lobe pattern with compact yet interleaved representations across latent steps, a sign of convergent organization where the trajectory remains bounded throughout the chain. In the SIM-CoT setting, the structure becomes more compact and the temporal separation between latent steps reduces further, indicating that the supervision tightens the convergence onto a single region.
\\
For the COCONUT method, the vanilla CoT setting shows each latent step occupying a distinct region of the manifold, with the beginning tokens (purple) and end tokens (yellow/green) positioned in close proximity, a possible indication of shortcut pathways where the model bridges early and late representations directly. Whereas, in the SIM-CoT setting, the distinct-region structure is preserved but the proximity between beginning and end tokens reduces, showing that the supervision retains the regional separation while reducing the shortcut tendency.

\begin{figure}[H]
\centering
\begin{minipage}{0.30\linewidth}
\centering
\includegraphics[width=\linewidth,trim={0 0 0 1cm},clip]{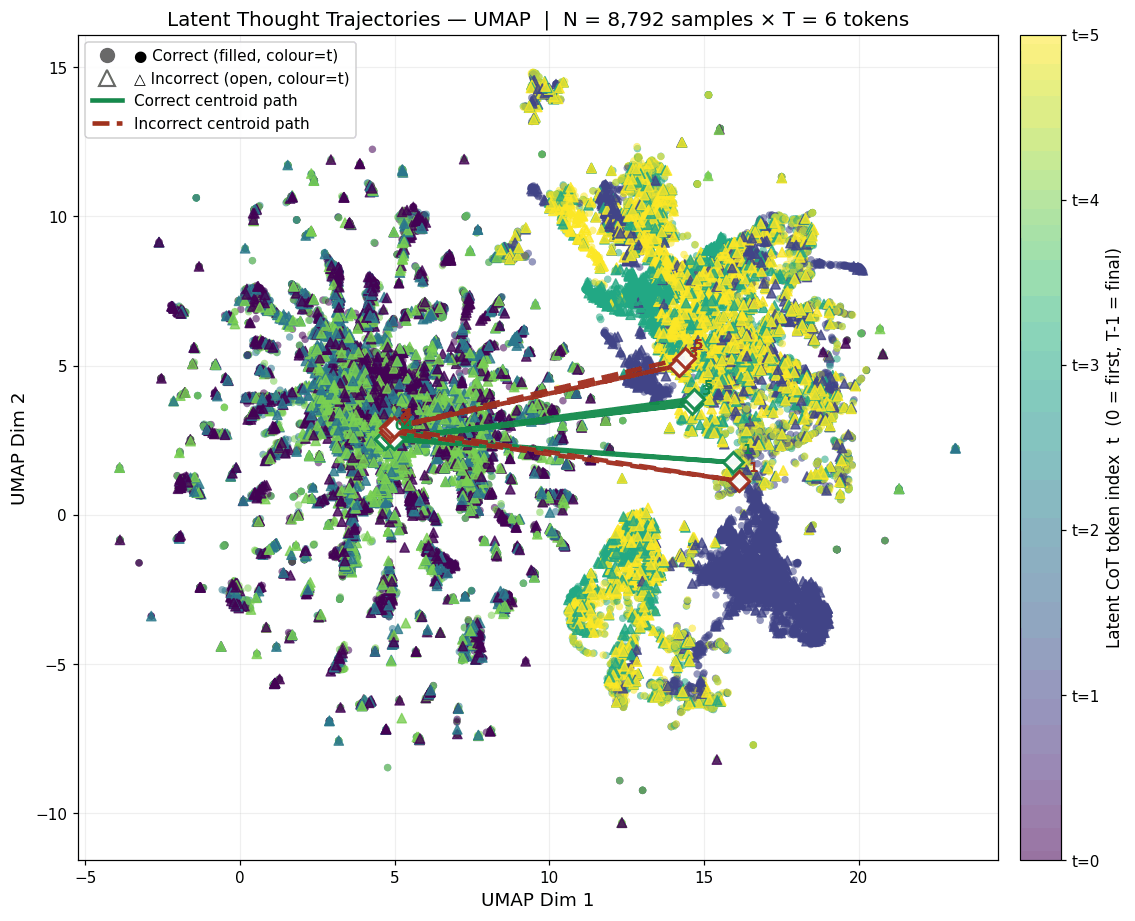}
\end{minipage}
\hspace{0.02\linewidth}
\begin{minipage}{0.30\linewidth}
\centering
\includegraphics[width=\linewidth,trim={0 0 0 1cm},clip]{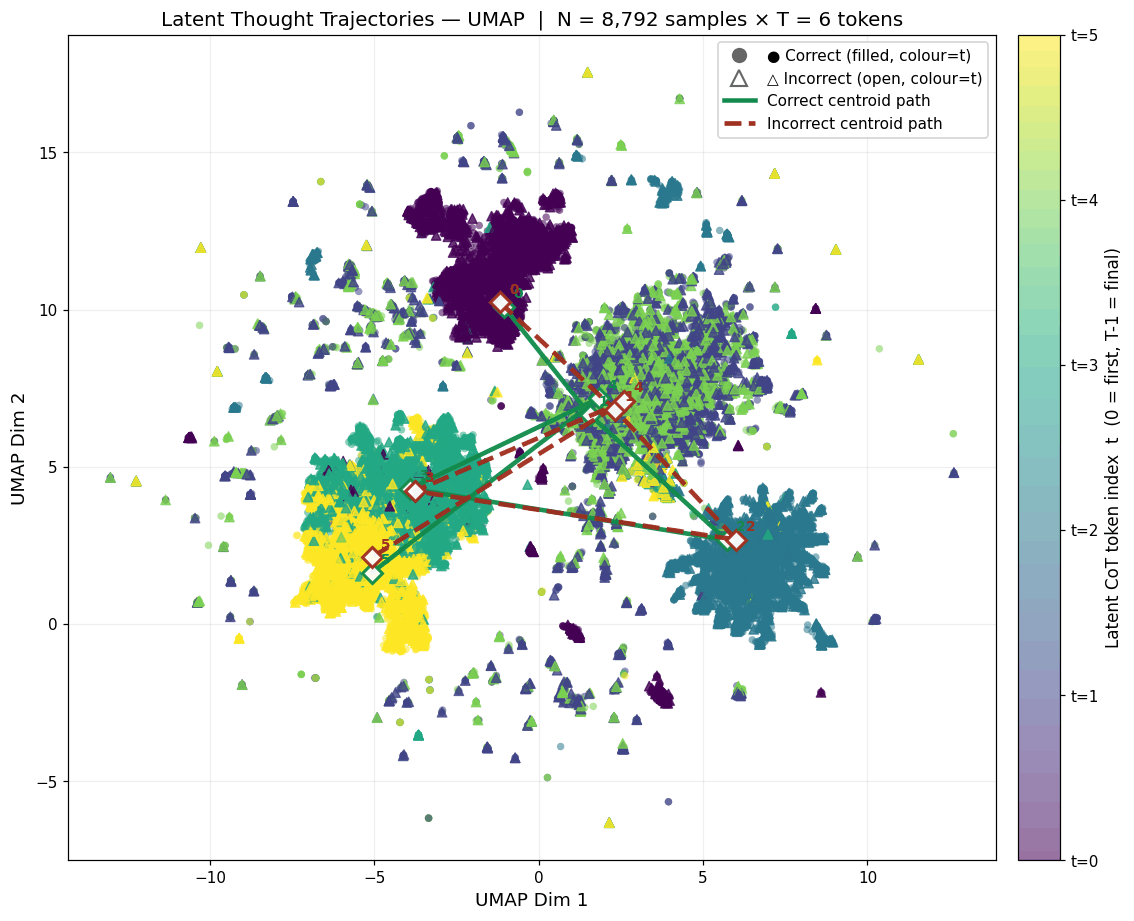}
\end{minipage}
\caption{\textbf{UMAP trajectory projections across latent steps for methods (CODI (left) and COCONUT(right))}: COCONUT shows each latent step occupying distinct regions, reinforcing step-specific reasoning roles. CODI forms two clusters — middle tokens diffused broadly, while beginning and end tokens remain densely contained. (Number of samples = 8,792 samples, 6 tokens)
}
\label{fig:traj_umap_gsm8k}
\end{figure}

\paragraph{UMAP} (as seen in Figure~\ref{fig:traj_umap_gsm8k}, right) projects the latent trajectory while preserving local neighborhood relationships and balancing global structure, showing how latent representations cluster by reasoning step.
\\
For the CODI method, the vanilla CoT setting shows two main clusters: middle tokens ($3^{rd}$ and $4^{th}$) scattered across one cluster while beginning and end tokens are contained in the other, a sign of role separation where mid-chain steps explore broadly and terminal steps remain bounded. In the SIM-CoT setting, the cluster boundaries tighten and the middle-token spread reduces, indicating that the supervision concentrates the latent representations more uniformly across steps.
\\
For the COCONUT method, the vanilla CoT setting shows each latent step occupying a distinct region of the embedding space, a sign that latent representations carry step-specific roles in the reasoning process. Whereas, in the SIM-CoT setting, the step-wise regional separation is preserved with tighter cluster boundaries per step, showing that the supervision sharpens the role separation without changing the underlying organization.

We also show the t-SNE and PCA plots for the above methods (GSM8k + Vanilla CoT setting) in the Appendix \ref{app:tsne} and \ref{app:pca} respectively.

\section{Discussion}
We present our findings based on the above analysis as follows,
\\\\
\textbf{Dynamics of CODI and COCONUT exhibit different latent stability types} - Lyapunov metric analysis (see Figure \ref{fig:lyapunovsensitivity_2x2}) highlights CODI behaves as a \textit{stable attractor} type, where latent representations converge as timestep approaches infinity, which is consistent with its two lobe pattern (see Figure \ref{fig:traj_phate_gsm8k} (left)). COCONUT, in contrast, exhibits \textit{unstable expanding} dynamics, where it starts with stable latent space and diverges as time progresses reflecting its butterfly pattern (see Figure \ref{fig:traj_dmd_gsm8k} (right)).

\textbf{SIM-CoT shows better latent stability than Vanilla Latent COT paradigm} - From a dynamical perspective, we see how SIM-CoT version of CODI and COCONUT show more stable behavior than their vanilla CoT counterparts in Step-to-Step change (Figure \ref{fig:step2stepchange_2x2}) and Lyapunov stability (Figure \ref{fig:lyapunovsensitivity_2x2}).
\\\\
\textbf{Latent geometry for CODI and COCONUT mirrors the training objective} - In Figure \ref{fig:traj_umap_gsm8k}, COCONUT exhibits more coherent and well-separated latent CoT structure, consistent with its curriculum training which transitions from explicit to implicit tokens. By contrast, CODI shows more diffuse latent trajectories, consistent with its training objective centered on distillation rather than enforcing structure across individual latent CoT tokens.
\\\\
\textbf{Underlying cognitive mechanisms for CODI and COCONUT are fundamentally different}
- Based on the latent representation structures seen in DMD plots, we notice the following,
\begin{itemize}
    \item COCONUT follows a \textit{computation} strategy where it \textbf{persists} all reasoning trajectories simultaneously (denoted by the spread seen in later stages) and continues to expand on it.
    \item CODI follows a \textit{classification} strategy where all the reasoning trajectories \textbf{converge to two modes} and the reasoning process moves between these modes before finalizing the solution. 
\end{itemize}
Overall, the two methods solve problems through fundamentally opposite dynamics. In COCONUT, doing more work in latent space is associated with getting the answer right. In CODI, shorter and simpler paths are the ones
that lead to correct answers.

\section{Conclusion}
\label{sec:conclusion}
This paper provides an insight into the  dynamics for the latent CoT reasoning methods. Leveraging an analytical framework grounded in dynamical systems, we demonstrate how these methods differ fundamentally in their stability profiles, training procedures, and the cognitive strategies used in problem solving. We show that interpreting latent CoT tokens as dynamical reasoning states provides the community with another axis of interpretability for further improvements in this space. 
\\
Our experiments mainly focus on the GSM8k dataset and the GPT-2 Small model. We also focus mainly on the CODI and COCONUT latent CoT reasoning methods, and additionally use the SIM-CoT training paradigm to show how are analysis is paradigm agnostic in nature.
\\
Future works should investigate on the stability types for CODI and COCONUT on long CoT chains, and whether
interventions in latent space can improve
downstream performance. Other directions include extending this framework to larger reasoning models like Llama, Deepseek-distill-Llama3-8b, etc. and other datasets like MATH and Strategy-QA.

\section*{Impact Statement}
Our framework establishes a trajectory-based analytical paradigm grounded in dynamical systems theory to formalize the interpretability of latent reasoning processes. This research facilitates the interpretability of reasoning dynamics in latent CoT reasoning technique, enabling the identification of dynamical structures and stability modes in explicit CoT. These findings offer a robust foundation for the deployment of resource-efficient latent reasoning architectures.

\FloatBarrier
\bibliography{fg_icml_example_paper}
\bibliographystyle{icml2026_fogen}

\newpage
\appendix
\onecolumn
\section{Appendix}
\label{app:gsm8k_plots}

\subsection{Hyperparameters}
\subsubsection{Dimensionality Reduction Hyperparameters}

All reduction methods project to $k = 2$ and $k = 3$ dimensions for visualization.
The following hyperparameters are fixed across all runs:
\begin{itemize}

    \item \textbf{t-SNE:} Perplexity $=5.0$ (auto-adjusted to $\min\!\bigl(5.0,\,\max(2.0,\,(N{-}1)/3)\bigr)$
for small batches).

    \item \textbf{UMAP:} $n_{\text{neighbors}} = 5$,
    $\text{min\_dist} = 0.1$, Euclidean metric.

    \item \textbf{DMD:} Full-rank SVD
    ($\text{svd\_rank} = -1$, effective rank $= \min(D,\,T{-}1) = 5$
    for $T{=}6$ step trajectories); batched GPU computation across all $N$
    samples simultaneously.
    Eigenvalue magnitudes $|\lambda|$ classify modes as stable
    ($|\lambda| < 1$) or unstable ($|\lambda| > 1$).

    \item \textbf{PHATE:} $k$-nearest neighbors $k = 5$,
    diffusion time $t = \text{auto}$.
\end{itemize}

\subsubsection{Perturbation Stability Hyperparameters}

Perturbation experiments inject Gaussian noise directly into input
embeddings and re-run inference.
Noise standard deviation $\sigma = 0.01$ and $n = 3$ independent
perturbation runs per sample are used.
Divergence is measured as the mean $\ell_2$ distance between clean and
perturbed trajectories at each step, alongside a scale-invariant relative
divergence $\lVert z_t^{\text{perturbed}} - z_t\rVert / \lVert z_t \rVert$.

\subsubsection{Metric Plot Conventions}

All metric plots display mean curves $\pm$ one standard deviation across
samples, split by prediction correctness (correct and incorrect).
In trajectory plots, each point represents one sample colored by latent step index $t$; prediction correctness is encoded by marker shape (filled circle = correct, open triangle = incorrect), with mean centroid paths overlaid.

\subsection{Additional Trajectory Visualizations}
\subsubsection{PCA Projections}
\label{app:pca}

Linear PCA projects the latent trajectory tensor onto the two directions of greatest
variance. Color encodes the latent step index from $t{=}0$ (purple) to $t{=}5$ (yellow);
filled circles mark correct outputs and open triangles mark incorrect outputs.
Solid green and dashed red lines show the correct and incorrect centroid paths respectively.

\begin{figure}[H]
\centering
\begin{minipage}{0.48\textwidth}
    \centering
    \includegraphics[width=\linewidth,trim={0 0 0 2.9cm},clip]{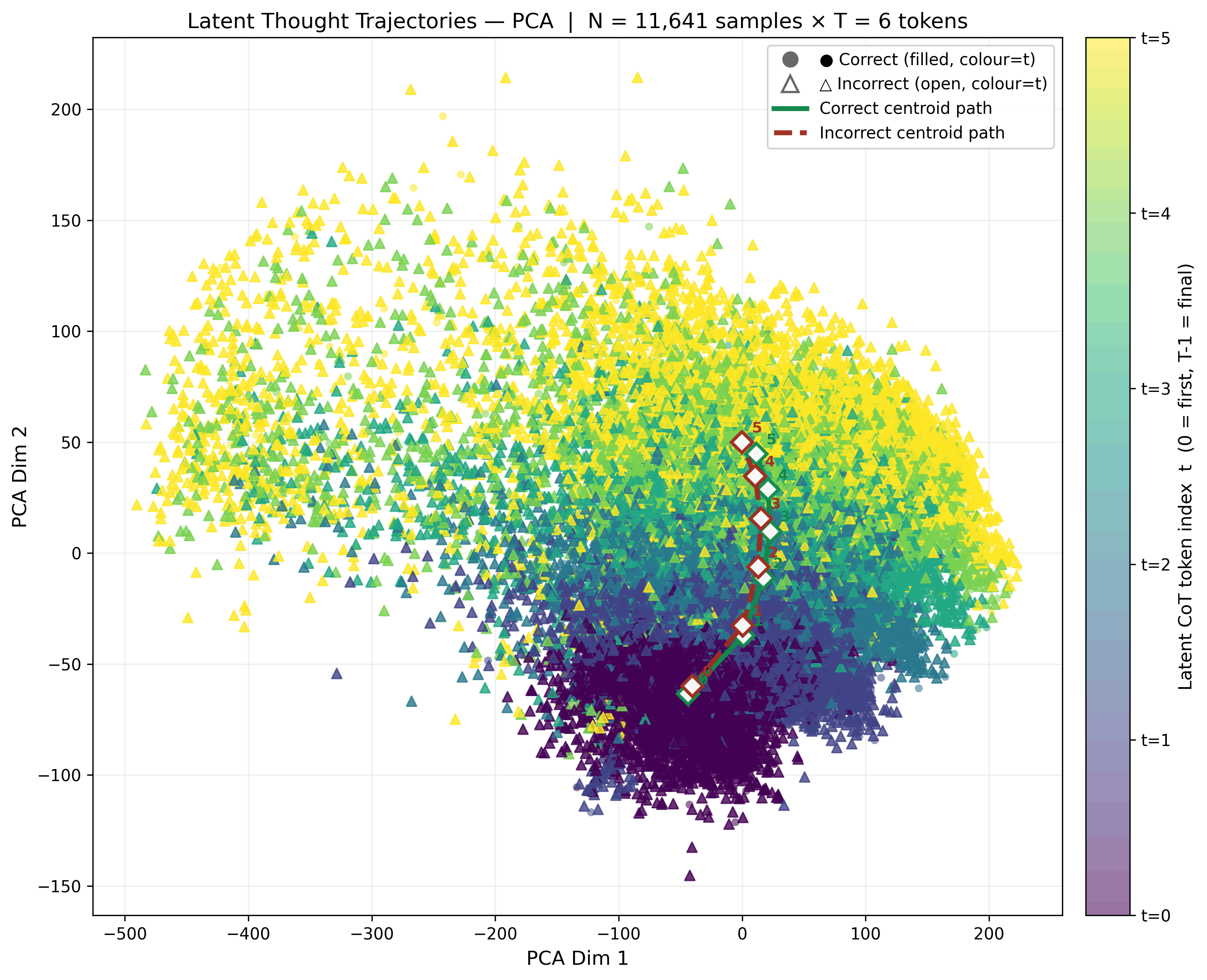}
    \subcaption{COCONUT · GSM8K}
\end{minipage}
\hfill
\begin{minipage}{0.48\textwidth}
    \centering
    \includegraphics[width=\linewidth,trim={0 0 0 2.9cm},clip]{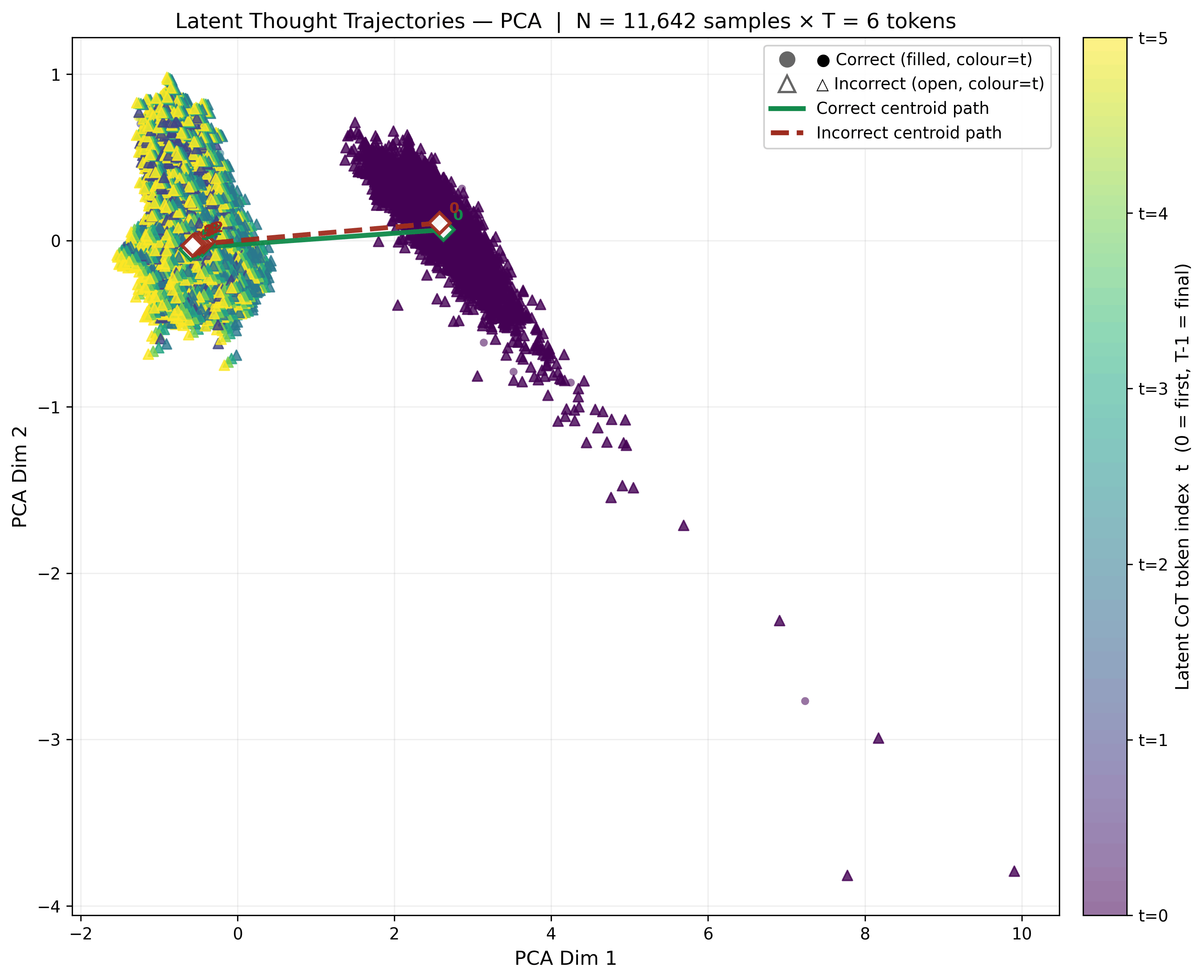}
    \subcaption{CODI · GSM8K}
\end{minipage}
\caption{PCA projections of latent trajectories for COCONUT (left) and CODI (right).}
\label{fig:traj_pca}
\end{figure}

For COCONUT, the point cloud splits bimodally along PC2 (the vertical axis): late steps
$t{=}4$ and $5$ occupy an upper lobe above PC2 $> 0$, while early steps $t{=}0$ and $1$
occupy a lower lobe below PC2 $< {-}10$. Mid-steps $t{=}2$ and $3$ bridge the two lobes.
Both centroid paths trace nearly identical zig-zag routes through this structure, sitting on top
of one another throughout the chain. PC2 is acting as a temporal axis: the principal directions
of variance encode where in the chain a vector lives, not whether the chain succeeds. Linear
PCA cannot separate correct from incorrect trajectories on this model.

For CODI, the bimodal structure is preserved but rotated: the split is along PC1 (the
horizontal axis) rather than PC2. Late steps $t{=}3$ to $5$ form a right lobe at PC1 $> 30$,
and early steps $t{=}0$ and $1$ form a left lobe at PC1 $< {-}30$, with a quieter central gap
between them. Centroid paths sweep horizontally from left to right. The two-lobe horizontal
axis in CODI mirrors COCONUT's two-lobe vertical axis, but the orientation reflects a
different dominant mode structure. In both models, the principal directions of variance encode
reasoning phase rather than problem type.

\subsubsection{t-SNE Projections}
\label{app:tsne}

t-SNE preserves local neighborhood structure and is used here to assess whether correct
and incorrect trajectories occupy distinct regions at each step. Unlike PCA, t-SNE is
nonlinear and does not preserve global distances.

\begin{figure}[H]
\centering
\begin{minipage}{0.48\textwidth}
    \centering
    \includegraphics[width=\linewidth,trim={0 0 0 2.8cm},clip]{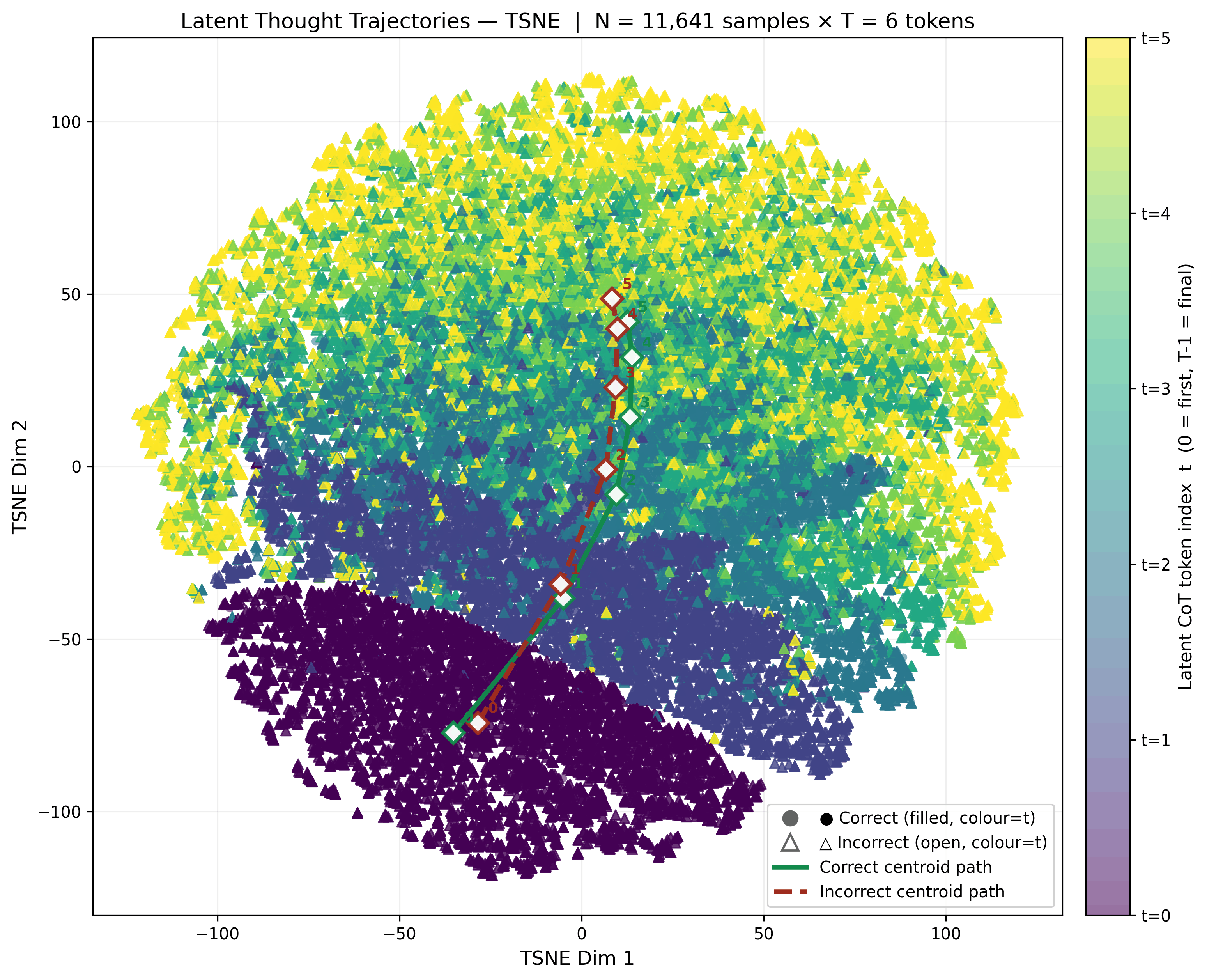}
    \subcaption{COCONUT · GSM8K}
\end{minipage}
\hfill
\begin{minipage}{0.48\textwidth}
    \centering
    \includegraphics[width=\linewidth,trim={0 0 0 2.8cm},clip]{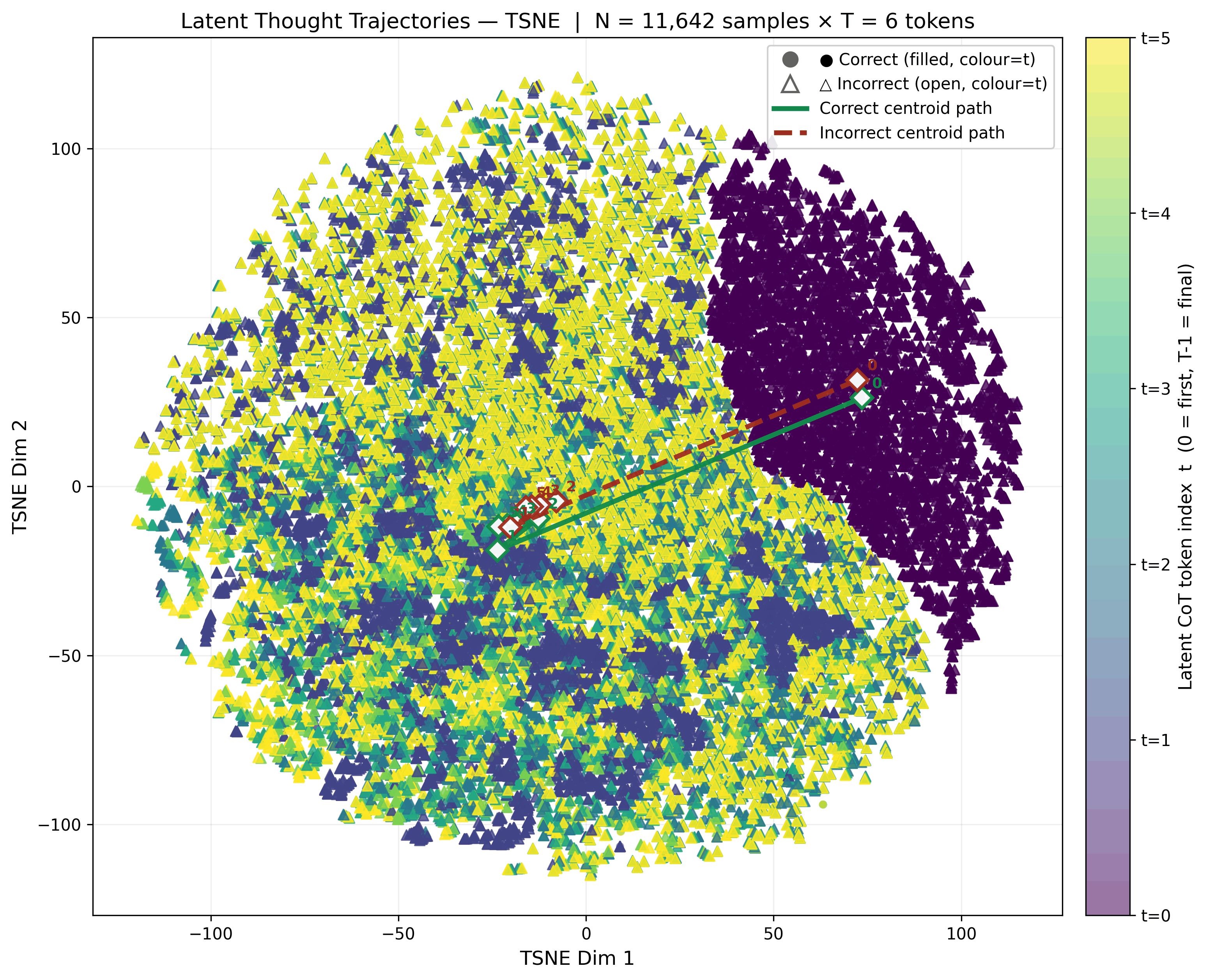}
    \subcaption{CODI · GSM8K}
\end{minipage}
\caption{t-SNE projections of latent trajectories for COCONUT (left) and CODI (right).}
\label{fig:traj_tsne}
\end{figure}

For COCONUT, the projection produces a roughly disc-shaped cloud spanning approximately
${\pm}100$ in both dimensions. A pronounced yellow island at $t{=}5$ sits in the upper center
of the disc, indicating that late-step representations collapse into a single localized attractor
region. Early steps ($t{=}0$) scatter across the periphery in distributed pockets. Both centroid
paths walk overlapping zig-zag routes through the dense middle of the disc. The correctness
signal in t-SNE coordinates is weak. The plot is best read as evidence that the late chain
commits to a small region while the early chain explores broadly — the local-neighborhood
signature of the same compression visible in the UMAP and DMD projections in the main paper.

For CODI, the disc shape is similar in overall envelope but the internal structure is more
diffuse: sub-clusters bleed into each other rather than forming sharp islands. Yellow $t{=}5$
vectors appear in scattered local pockets without a single clean breakaway region. A dark
purple concentration of early steps sits in the upper portion of the disc; mid-step teal forms a
diffuse band through the middle. Centroid paths run as short, heavily overlapping trajectories
through the dense center. The correctness signal is weak, as in COCONUT. This diffuseness
is consistent with CODI's distillation procedure flattening the latent geometry: the model has
many equally important local structures rather than a few dominant attractor basins.


\subsubsection{3D Qualitative Trajectory plots of Vanilla CODI and COCONUT with GSM8K}
\label{3d-plot-gsm8k-codi-coconut}

\begin{figure}[H]
\centering
\begin{minipage}{0.48\linewidth}
\centering
\includegraphics[width=\linewidth,trim={0 0 0 2.9cm},clip]{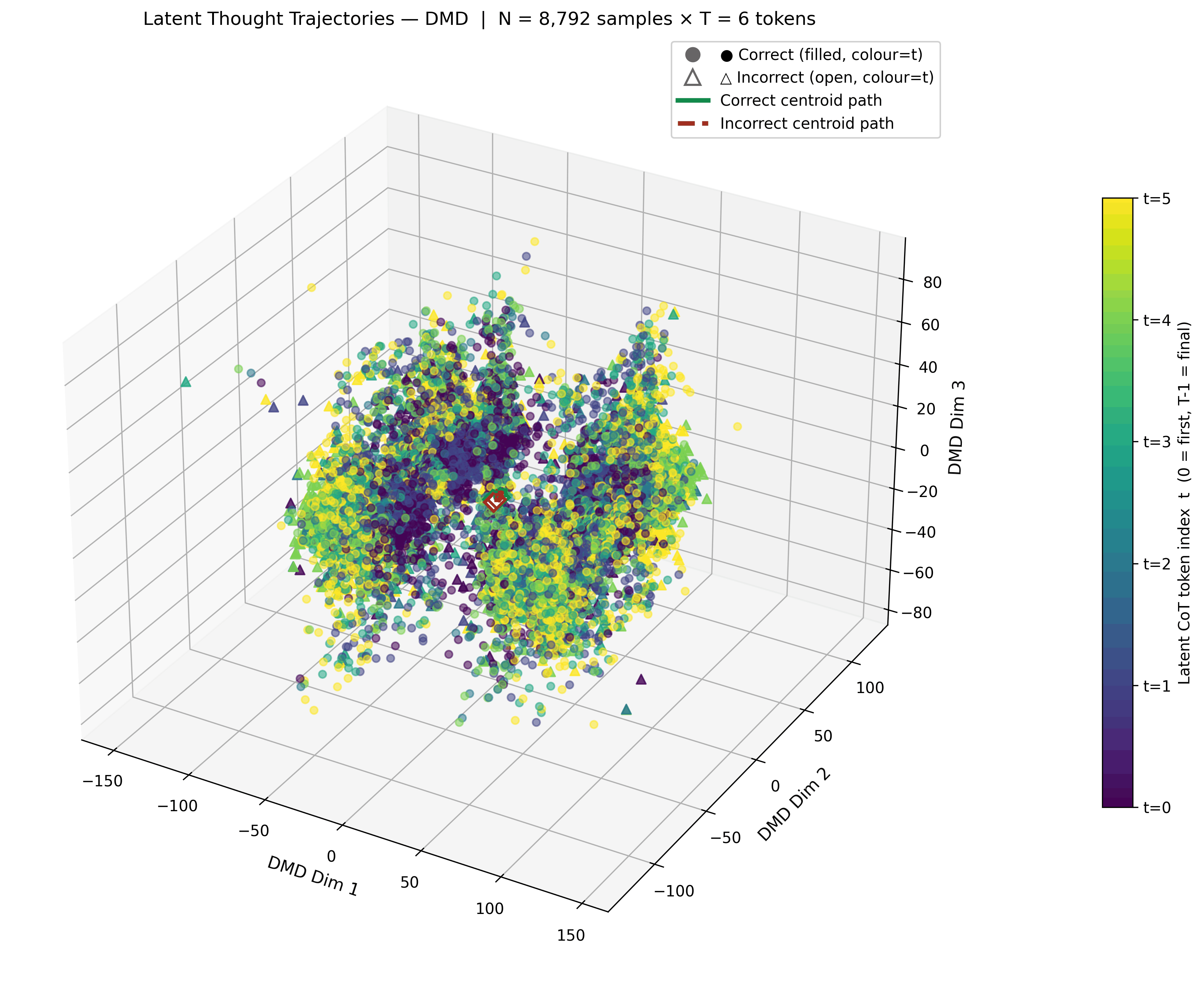}
\end{minipage}
\hfill
\begin{minipage}{0.48\linewidth}
\centering
\includegraphics[width=\linewidth,trim={0 0 0 2.9cm},clip]{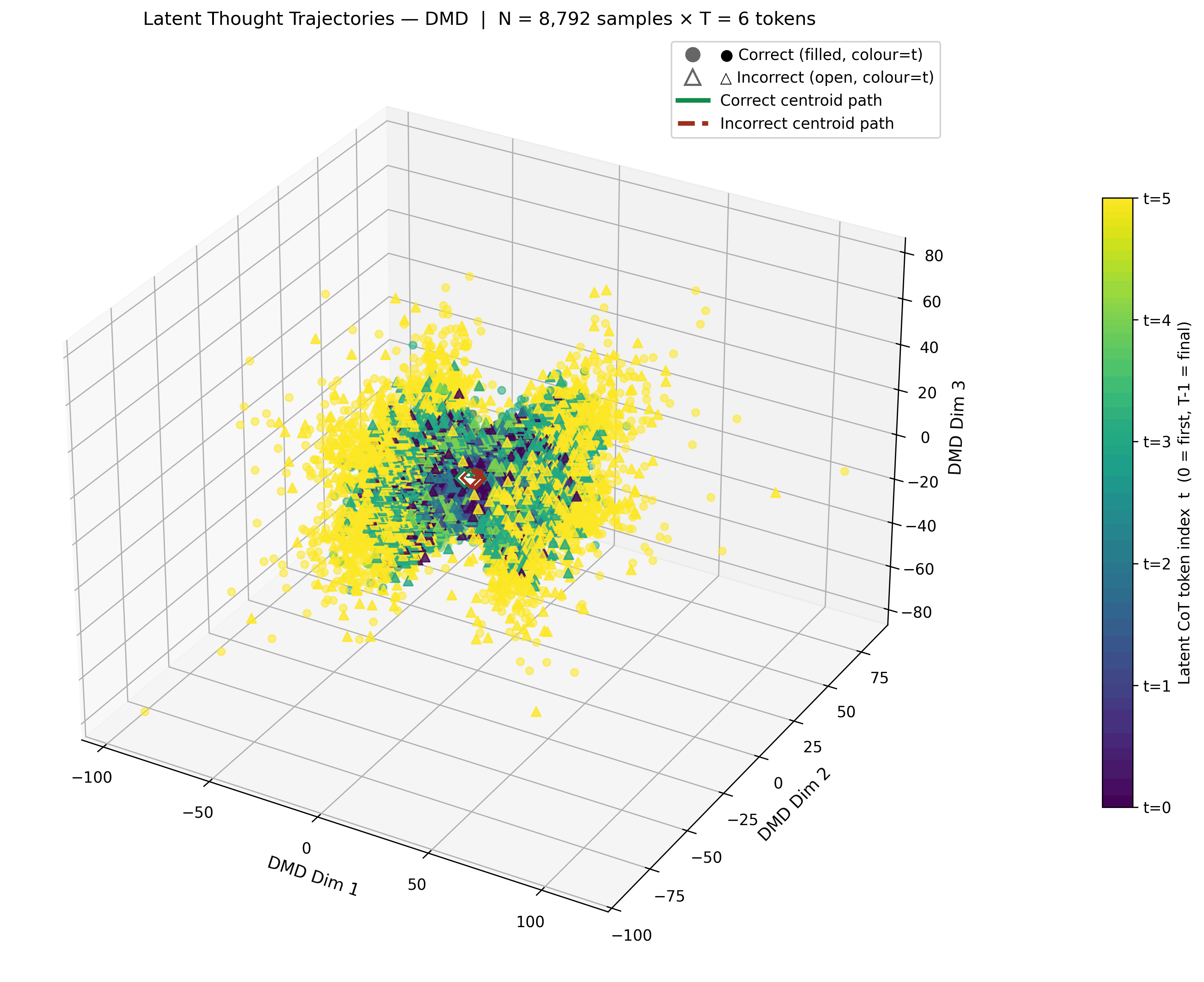}

\end{minipage}
\caption{3D DMD trajectory projections across latent steps for CODI (left) and COCONUT(right). (Number of samples = 8,792 samples, 6 tokens)}
\label{fig:3d_traj_dmd_gsm8k}
\end{figure}

\paragraph{DMD}
For COCONUT, the latent states start near the center for $t{=}0$ and spread outward by $t{=}5$ showing that the model starts diverging through latent space at later reasoning steps proving its "chaotic" behaviour as described by direction consistency (Figure~\ref{fig:3d_traj_dmd_gsm8k}, right).

Unlike COCONUT, CODI shows a unclear similar divergence in latent steps as they increase (Figure~\ref{fig:3d_traj_dmd_gsm8k}, left).

\begin{figure}[H]
\centering
\begin{minipage}{0.48\linewidth}
\centering
\includegraphics[width=\linewidth,trim={0 0 0 3cm},clip]{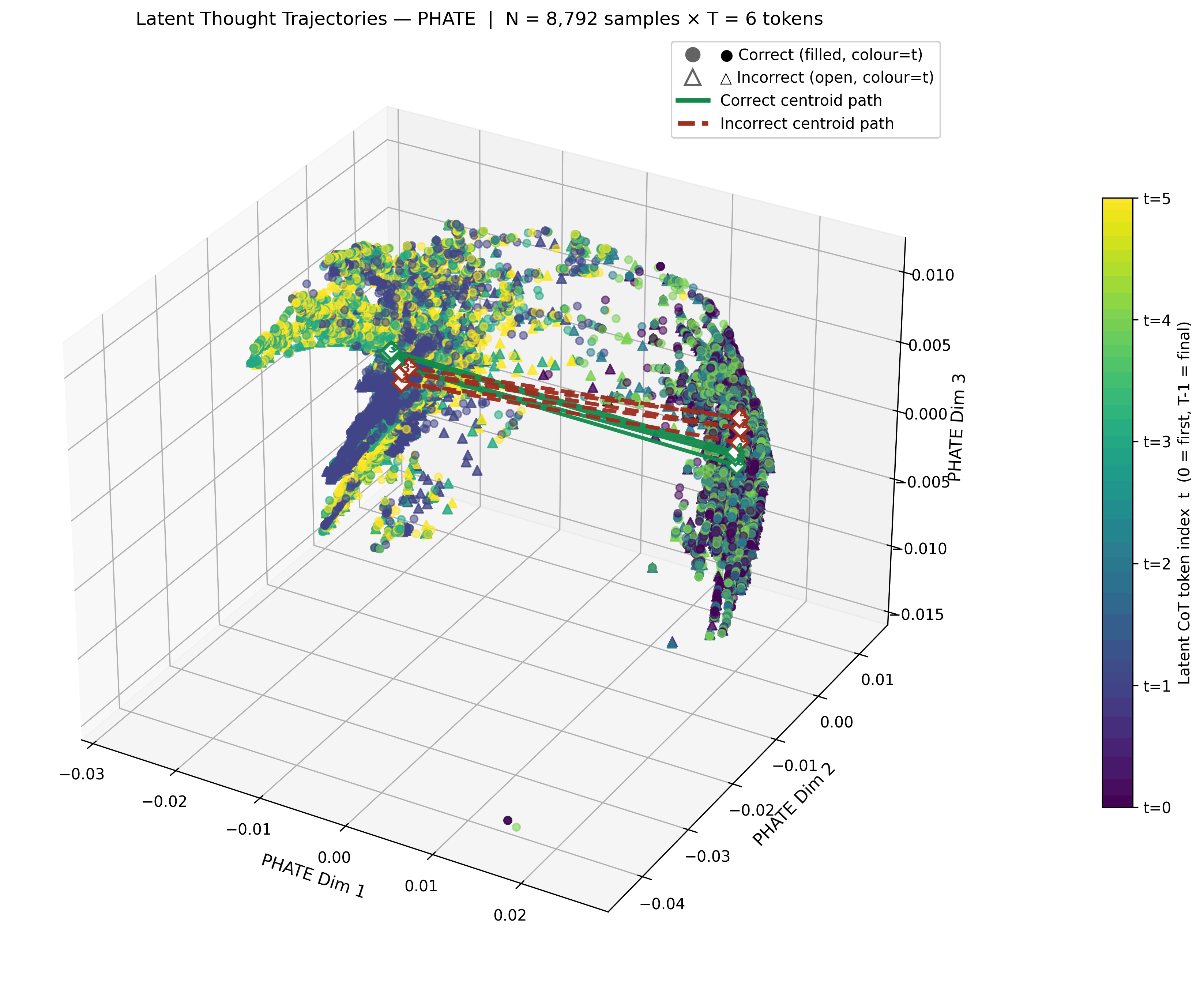}
\end{minipage}
\hfill
\begin{minipage}{0.48\linewidth}
\centering
\includegraphics[width=\linewidth,trim={0 0 0 3cm},clip]{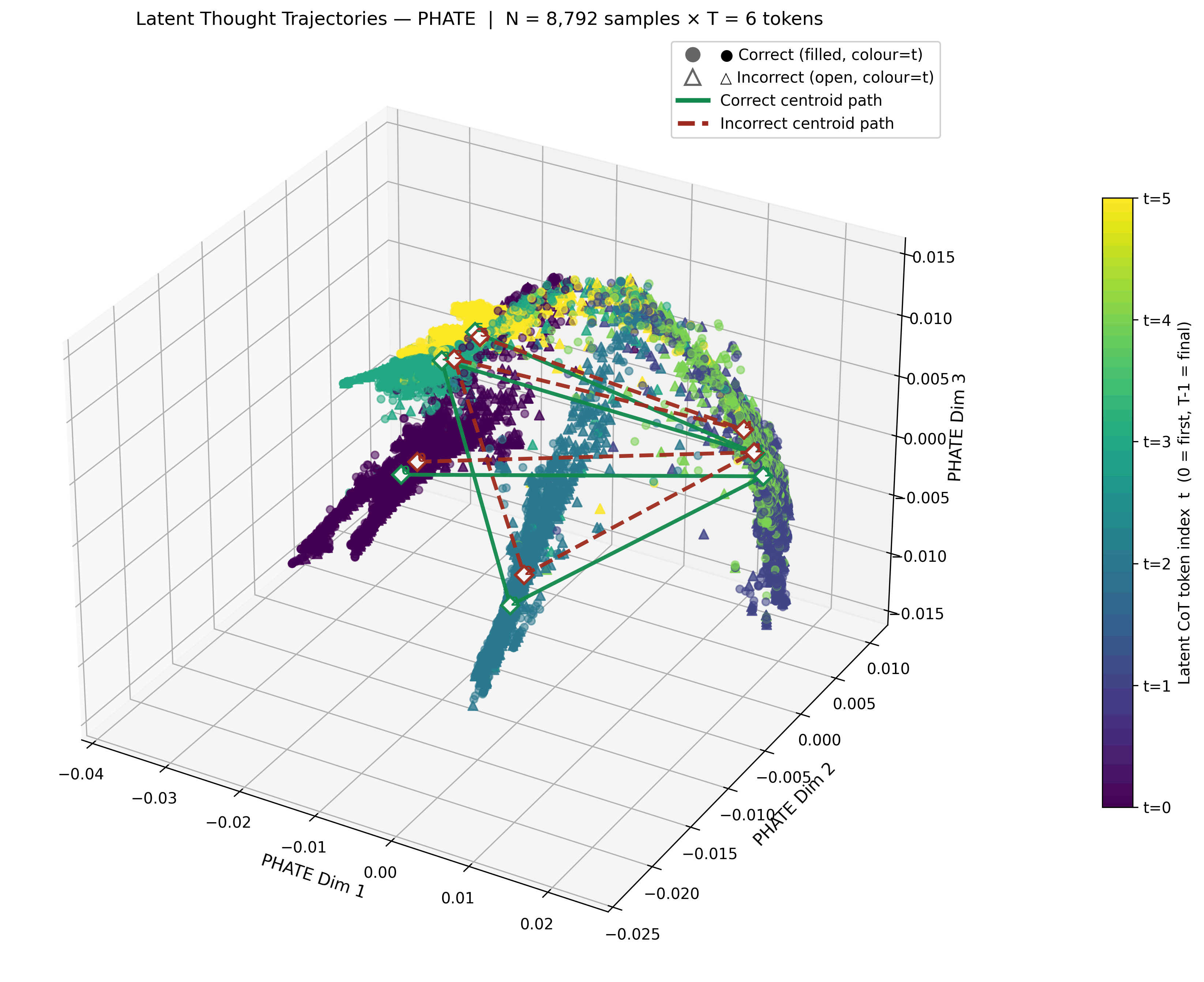}
\end{minipage}
\caption{3D PHATE trajectory projections across latent steps for CODI (left) and COCONUT(right).(Number of samples = 8,792 samples, 6 tokens)}
\label{fig:3d_traj_phate_gsm8k}
\end{figure}

\paragraph{PHATE}
We can see from the PHATE projections of COCONUT that each latent step explores different latent regions demonstrating their distinct role in the reasoning process (Figure~\ref{fig:3d_traj_phate_gsm8k}, right).

Unlike COCONUT, CODI shows a unclear similar clustering between latent steps (Figure~\ref{fig:3d_traj_phate_gsm8k}, left).

\subsubsection{3D Qualitative Trajectory Plots of SIM-CoT CODI and COCONUT with GSM8K}

\paragraph{DMD}
DMD Unlike Vanilla CODI (Figure 10, left), CODI under SIM-CoT paradigm
(Project Page, Section~S.1\footnote{\url{https://sabariiyyappan.github.io/Latent-CoT/}}) shows that the latent states start near the center for $t{=}0$ and spread outward by $t{=}5$ showing that the model starts diverging through latent space at later reasoning steps like Vanilla COCONUT (Figure~\ref{fig:3d_traj_dmd_gsm8k}, right) and SIM-CoT COCONUT (Project Page, Section~S.1).

\paragraph{PHATE}
We can see from the COCONUT PHATE projections that each latent step explores different latent regions, demonstrating their distinct role in the reasoning process (Project Page, Section~S.1\footnote{\url{https://sabariiyyappan.github.io/Latent-CoT/}}).

Unlike COCONUT, CODI PHATE shows a unclear similar clustering between latent steps (Project Page, Section~S.1).

\subsection{Metric Plots}
\subsubsection{Arc Length}
\label{sec:A.3.1_arc_len}
Figures 12-15 show the arc length distributions for CODI and COCONUT under Vanilla and SIM-CoT\footnote{SIM-CoT is described in Section~S.2 of the project page: https://sabariiyyappan.github.io/Latent-CoT/} settings on GSM8K ($N = 8{,}792$ each).Each figure shows three panels: (a) the arc-length histogram split by correct (green) and incorrect (red) predictions with mean lines, (b) a boxplot of arc length by correctness, and (c) a per-concept strip plot with mean $\pm 1\sigma$ across the seven GSM8K concept categories. For both methods and both training paradigms, correct predictions show slightly longer arc lengths than incorrect ones, with CODI's distribution shifted to a higher absolute range than COCONUT's; SIM-CoT compresses CODI's distribution while leaving COCONUT's largely unchanged, indicating that the supervision reduces CODI's reasoning effort without altering COCONUT's.

\begin{figure}[h]
    \centering
    \includegraphics[width=1\linewidth,trim={0 0 0 3cm},clip]{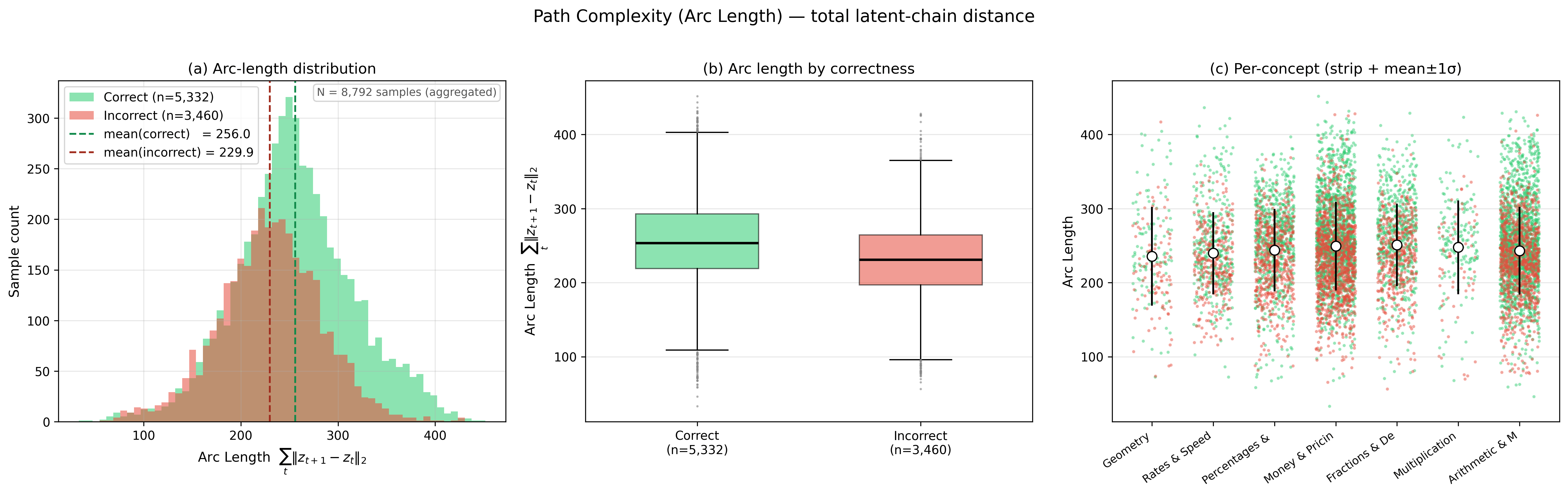}
    \caption{Arc length distribution for CODI on GSM8K (Vanilla setting,$N = 8{,}792$).}
    \label{fig:arclength-codi-vanilla}
\end{figure}
\begin{figure}[h]
    \centering
    \includegraphics[width=1\linewidth,trim={0 0 0 3cm},clip]{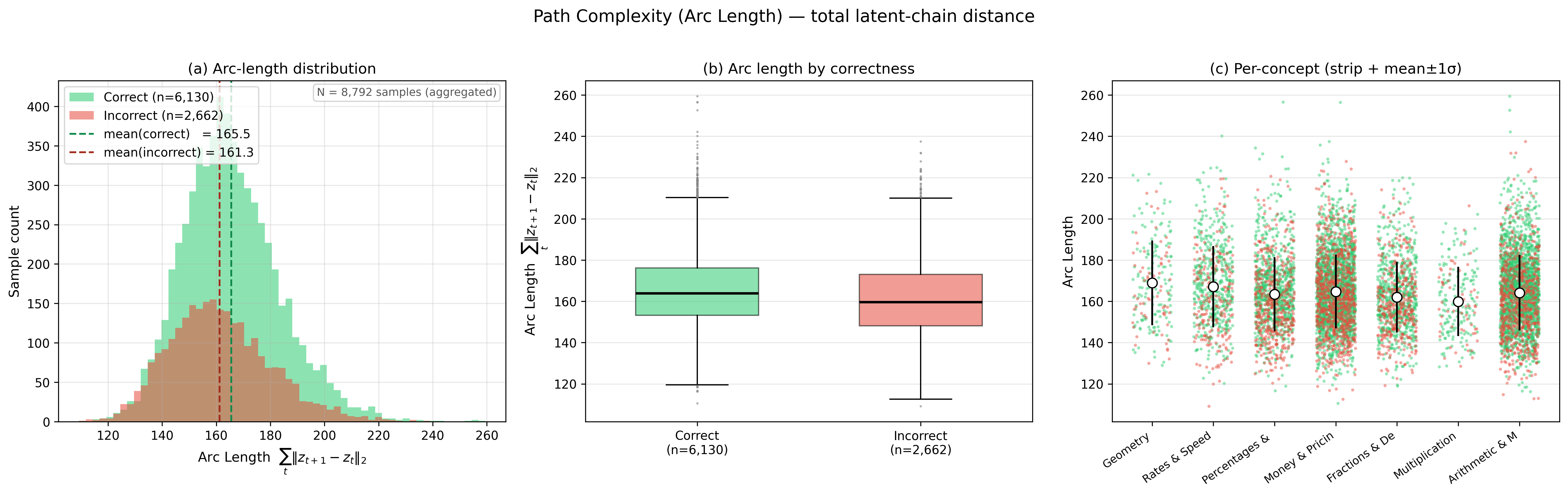}
    \caption{Arc length distribution for COCONUT on GSM8K (Vanilla setting,$N = 8{,}792$).}
    \label{fig:arclength-coconut-vanilla}
\end{figure}



\subsubsection{Perturbation stability}
\label{sec:A.3.2_perturb}
\begin{figure}[h]
    \includegraphics[width=1\linewidth,trim={0 0 0 5.5cm},clip]{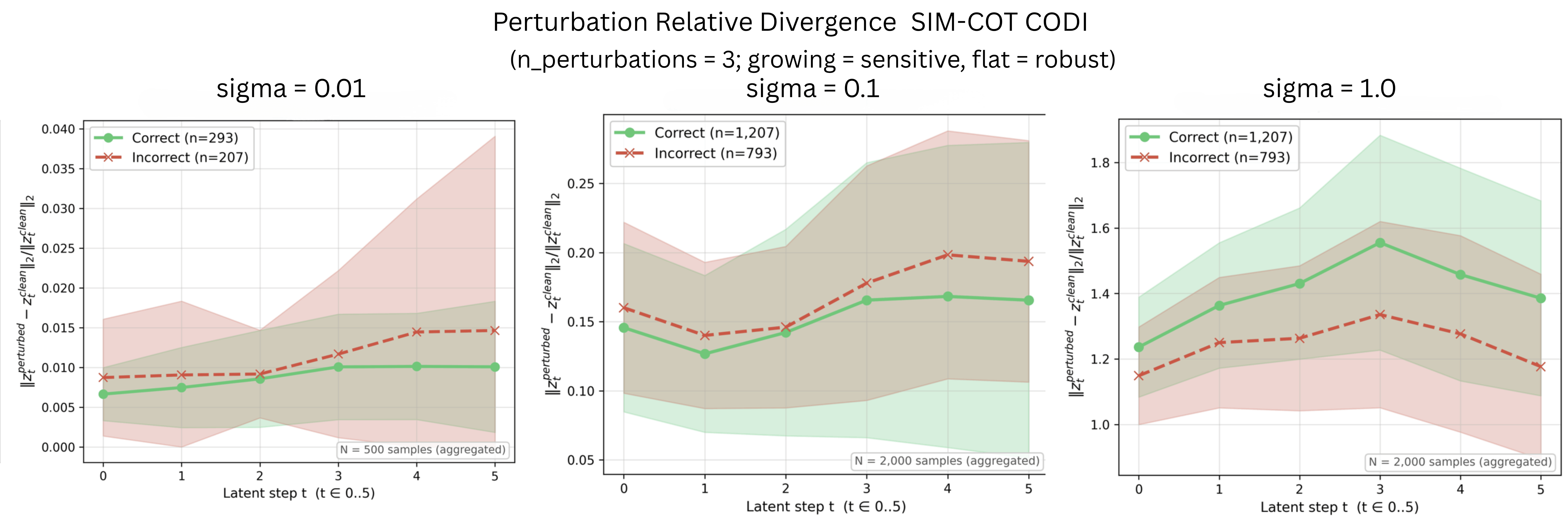}
    \caption{Perturbation relative divergence for SIM-CoT CODI on GSM8K at $\sigma = 0.01$ (left), $\sigma = 0.1$ (middle),and $\sigma = 1.0$ (right).}
    \label{fig:perturb-simcot-codi}
\end{figure}

\begin{figure}[h]
    \includegraphics[width=1\linewidth,trim={0 0 0 5.5cm},clip]{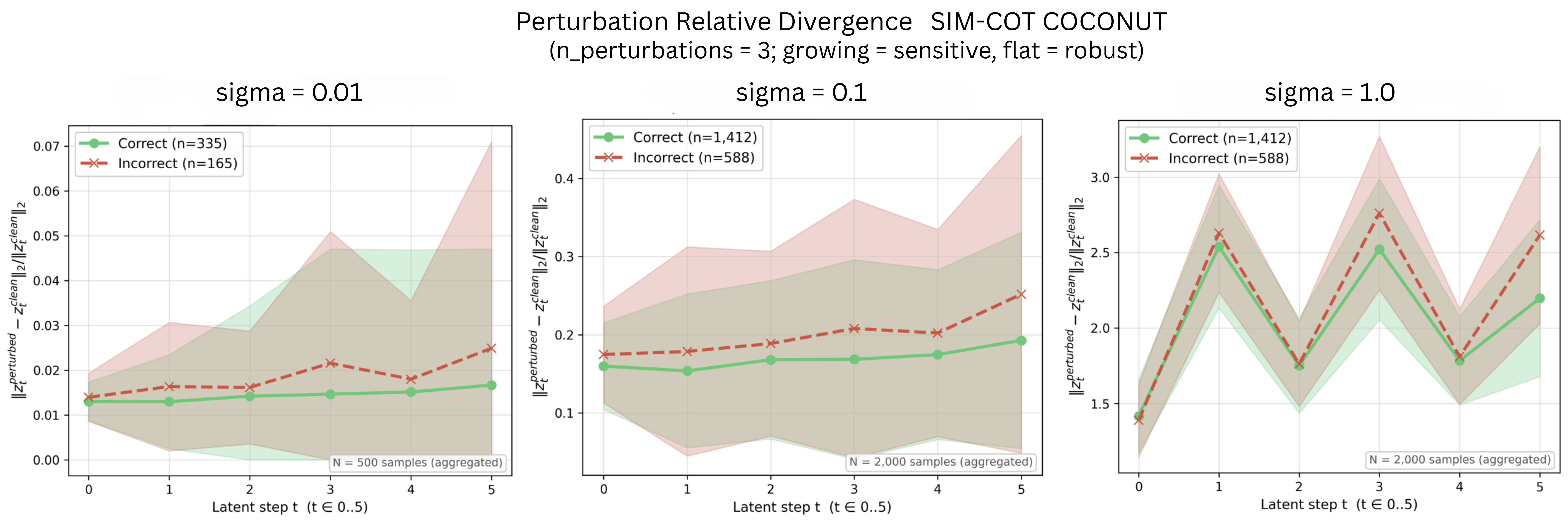}
    \caption{Perturbation relative divergence for SIM-CoT COCONUT on GSM8K at $\sigma = 0.01$ (left), $\sigma = 0.1$ (middle),and $\sigma = 1.0$ (right).}
    \label{fig:perturb-simcot-coconut}
\end{figure}

Figures~\ref{fig:perturb-simcot-codi} and~\ref{fig:perturb-simcot-coconut} show the perturbation relative divergence 
\[\|z_t^{\text{perturbed}} - z_t^{\text{clean}}\|_2 / \|z_t^{\text{clean}}\|_2\]
 for SIM-CoT CODI and SIM-CoT COCONUT on GSM8K across three Gaussian noise levels ($\sigma \in \{0.01, 0.1, 1.0\}$). Each panel shows correct (green) and incorrect (red) mean curves with $\pm 1$ standard-deviation bands across $n = 3$ perturbation runs per sample. The incorrect-above-correct ordering is preserved at $\sigma \in \{0.01, 0.1\}$ for both methods,a sign that uncertain reasoning paths are more sensitive to input-embedding perturbations; COCONUT's relative divergence sits consistently higher than CODI's at every $\sigma$, reflecting COCONUT's smaller clean-state magnitudes amplifying the proportional noise effect.


\subsubsection{Concept-wise Metric Plots}

Concept-wise metric plots are provided in Section~S.3 of the project page\footnote{\url{https://sabariiyyappan.github.io/Latent-CoT/\#supp-concept}}. The supplementary figures include step-to-step change, direction consistency, and Lyapunov sensitivity analyses for both CODI and COCONUT across the seven GSM8K math concept categories. Each panel shows correct (green) and incorrect (red) mean curves with standard deviation bands for a single concept. The per-concept profiles are consistent with the aggregated results presented in the main paper, indicating that the observed dynamics are primarily a property of the model architecture rather than the distribution of problem types.


\end{document}